%% file: root.tex
\documentclass[letterpaper, 10 pt, conference]{ieeeconf}  

\IEEEoverridecommandlockouts                              

\usepackage{cite}
\usepackage{amsmath,amssymb,amsfonts}
\usepackage{graphicx}
\usepackage{textcomp}
\usepackage{xcolor}

\def\BibTeX{{\rm B\kern-.05em{\sc i\kern-.025em b}\kern-.08em
    T\kern-.1667em\lower.7ex\hbox{E}\kern-.125emX}}

\input{doc/packages}
\input{doc/preamble}
\input{doc/math}

\input{doc/00-title}

\begin{document}

\maketitle
\copyrightnotice%
\input{doc/00-abstract}

\input{doc/01-introduction}
\input{doc/02-related-work}
\input{doc/03-foundations}
\input{doc/04-problem-formulation}
\input{doc/05-method}
\input{doc/06-evaluation}
\input{doc/07-conclusion}



%



\bibliographystyle{./IEEEtran.bst} 
\bibliography{./IEEEabrv, manual_references.bib}

\end{document}

%% file: doc/packages.tex
\usepackage{algorithm}
\usepackage{algorithmicx}  
\usepackage[noend]{algpseudocode}
\usepackage{amsfonts}
\usepackage{amsmath}
\usepackage{amsthm}
\usepackage{bm}
\usepackage{booktabs}
\usepackage{diagbox}
\usepackage{dsfont} 
\let\labelindent\relax  
\usepackage{enumitem}
\usepackage{etoolbox}
\usepackage{adjustbox}
\usepackage{graphicx}  
\usepackage{svg}
\usepackage{url}
\usepackage[breaklinks]{hyperref} 
\usepackage{lipsum}  
\usepackage{mathtools}
\usepackage{multirow}
\usepackage{nicefrac}
\usepackage[final]{pdfpages}
\usepackage{siunitx}
\usepackage[caption=false]{subfig}
\usepackage{tabu}
\usepackage{threeparttable}  
\usepackage[bottom]{footmisc}

\usepackage[textsize=small]{todonotes}
\usepackage{xcolor}
\usepackage{xfrac}
\usepackage{blindtext}
\usepackage{placeins}

\usepackage[normalem]{ulem}

\usepackage{pgfplots}
\pgfplotsset{compat=newest}
\usepgfplotslibrary{groupplots}
\usepgfplotslibrary{fillbetween}
\usepgfplotslibrary{colormaps}

\usetikzlibrary{pgfplots.statistics} 
\usetikzlibrary{pgfplots.groupplots}
\usetikzlibrary{arrows.meta}

\usepackage{mathabx}   

\usepackage{marvosym}  

%% file: doc/preamble.tex
\graphicspath{{img/}}
\hypersetup{colorlinks=false}

\definecolor{githubColor}{HTML}{2EA44F}

\definecolor{newGray}{HTML}{808080}

\definecolor{lightGray}{rgb}{0.7, 0.7, 0.7}%

\definecolor{backGray}{rgb}{0.3, 0.3, 0.3}%
\definecolor{matlabYellow}{rgb}{0.9290, 0.6940, 0.1250}%
\definecolor{matlabPurple}{rgb}{0.4940, 0.1840, 0.5560}%
\definecolor{matlabLBlue}{rgb}{0.3010, 0.7450, 0.9330}%
\definecolor{matlabGreen}{rgb}{0.4660, 0.6740, 0.1880}%
\definecolor{matlabRed}{rgb}{0.8500, 0.3250, 0.0980}%
\definecolor{matlabBlue}{rgb}{0, 0.4470, 0.7410}%
\definecolor{matlabDarkRed}{rgb}{0.6350 0.0780 0.1840}%

\definecolor{TAB_RED}{RGB}{214,39,40}
\definecolor{TAB_BLUE}{RGB}{31, 119, 180}
\definecolor{TAB_ORANGE}{RGB}{255, 127, 14}

 \pgfplotsset{
    colormap={custom_hot}{color(1cm)=(yellow); color(2cm)=(orange); color(3cm)=(red)}
} 

\DeclareMathOperator*{\argminA}{argmin} %

\pgfplotsset{
    trajectory_legend_img/.style={
        legend image code/.code={%
            \node[anchor=center, blue] at (-0.1cm,-0.06cm) {\textbullet};
            \node[anchor=center, cyan] at (-0.05cm,-0.06cm) {\textbullet};
            \node[anchor=center, green] at (0.0cm,-0.06cm) {\textbullet};
            \node[anchor=center, yellow] at (0.05cm,-0.06cm) {\textbullet};
            \node[anchor=center, red] at (0.1cm,-0.06cm) {\textbullet};
        }
    },
}

\pgfplotsset{
     compat/labels=pre 1.3 
}

\definecolor{colorCircle}{HTML}{0072BD}

\definecolor{colorRect}{HTML}{D95319}

\newcolumntype{O}[1]{S[detect-weight, mode=text, table-format=#1]}

\renewcommand{\bfseries}{\fontseries{b}\selectfont} 
\robustify\bfseries             
\newrobustcmd{\B}{\bfseries}

\newcommand\copyrighttext{\footnotesize \textcopyright~2025 IEEE. Personal use of this material is permitted. Permission from IEEE must be obtained for all other uses, in any current or future media, including reprinting/republishing this material for advertising or promotional purposes, creating new collective works, for resale or redistribution to servers or lists, or reuse of any copyrighted component of this work in other works.
}

\newcommand\copyrightnotice{%
    \begin{tikzpicture}[remember picture,overlay]%
 	\node[anchor=south, xshift=-0pt, yshift=20pt] at (current page.south)%
 	{\fbox{\parbox{\dimexpr\textwidth-\fboxsep-\fboxrule\relax}{\copyrighttext}}};%
 	\end{tikzpicture}%
}



%% file: doc/math.tex
\newtheoremstyle{tstyle}
  {}
  {}
  {\itshape}
  {}
  {\bfseries}
  {.}
  { }
  {\thmname{#1}\thmnumber{ #2}\thmnote{ (#3)}}%
\theoremstyle{tstyle}

\newcommand{\diag}{\operatorname{diag}}



%% file: doc/00-title.tex
\title{%
\LARGE \bf Dynamic Objective MPC for Motion Planning of Seamless Docking Maneuvers%
}

\author{Oliver Schumann, Michael Buchholz, and Klaus Dietmayer%
\thanks{This work was supported by the State Ministry of Economic Affairs, Labour and Tourism Baden-Württemberg (project U-Shift\,II, AZ\,3-433.62-DLR/60).}%
\thanks{All authors are with the Institute of Measurement, Control and Microtechnology, Ulm University, Albert-Einstein-Allee 41, 89081 Ulm, Germany {\tt\footnotesize \{firstname\}.\{lastname\}@uni-ulm.de}}%
}

%% file: doc/00-abstract.tex
\begin{abstract}
Automated vehicles and logistics robots must often position themselves in narrow environments with high precision in front of a specific target, such as a package or their charging station.
Often, these docking scenarios are solved in two steps: path following and rough positioning followed by a high-precision motion planning algorithm.
This can generate suboptimal trajectories caused by bad positioning in the first phase and, therefore, prolong the time it takes to reach the goal.
In this work, we propose a unified approach, which is based on a Model Predictive Control (MPC) that unifies the advantages of Model Predictive Contouring Control (MPCC) with a Cartesian MPC to reach a specific goal pose.
The paper's main contributions are 
the adaption of the dynamic weight allocation method to reach path ends and goal poses inside driving corridors, and the development of the so-called dynamic objective MPC.
The latter is an improvement of the dynamic weight allocation method, which can inherently switch state-dependent from an MPCC to a Cartesian MPC to solve the path-following problem and the high-precision positioning tasks independently of the location of the goal pose seamlessly by one algorithm.
This leads to foresighted, feasible, and safe motion plans, which can decrease the mission time and result in smoother trajectories.
\end{abstract}


%% file: doc/01-introduction.tex
\FloatBarrier
\section{Introduction}
\label{sec:intro}
\begin{figure}[t]
    \centering
    \input{plots/introduction/overview}%
    \caption{%
    Overview of the proposed methods.
    1. Precise positioning at switching points by dynamic weight allocation while being able to smooth the rest of the trajectory; 2. Seamless and safe planning to goal poses by dynamic objective allocation while respecting corridor constraints.
    }%
    \label{fig:overview}
\end{figure}
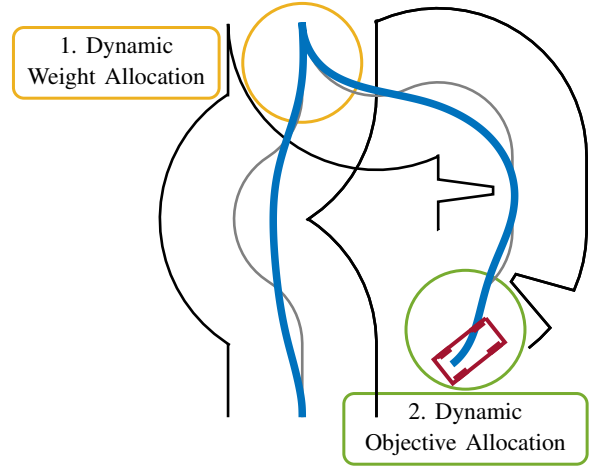%
Automated guided vehicles and other logistics robots are increasingly used in warehouses and similar environments \cite{warehouseReport}.
In these scenarios, the robots are faced with the task of finding, transporting, and dropping off different goods or containers in narrow and unstructured environments. 
Further, they regularity need to dock automatically to their charging station. 
In this work, we focus on the motion planning part of these use cases.
There already exist industry-level solutions, many of them using the motion planning algorithms of the Nav2 stack of Open Navigation LLC~\cite{macenski2020marathon2, macenski2023survey}.
These algorithms provide methods for motion planning in ROS2~\cite{macenski2022ros} and contain a docking mechanism for connecting robots and vehicles with charging stations or similar objects.
Generally, these approaches are often divided into two phases.
A rough positioning to a so-called staging pose followed by the detection of, for example, the charging station and the navigation towards it.
This workflow has two main disadvantages.
The robot stops at the staging pose to allow for a safe switch to another planning algorithm, which is often necessary.
This intermediate stop takes time, and it is unlikely that the staging pose is part of the optimal trajectory to the goal pose.
In addition to that, the controller used in the Nav2 stack to reach the final goal pose in the second phase is not able to handle non-holonomic vehicle dynamics.
Hence, if the goal pose is at an inconvenient position that cannot be reached physically from the staging pose, the controller will fail to reach it, which will trigger a retrial of the procedure starting at a new staging pose.

In this work, we mitigate these disadvantages by proposing a unified MPC-based approach that does not require switching between two algorithms but seamlessly completes the docking scenario.
The main contributions of this work are visualized in Fig.~\ref{fig:overview}, which are:%
\begin{enumerate}
    \item Dynamic weight allocation to precisely reach direction changes and path ends. This serves as the base for the last contribution, which is called
    \item Dynamic objective allocation: An MPCC that inherently changes its objective to transition to a pure cartesian MPC to reach specific goal poses.
\end{enumerate}
The source code of this paper will be published on the final submission at\footnote{\href{https://github.com/uulm-mrm/dynamic_objective_mpc}{github.com/uulm-mrm/dynamic\_objective\_mpc}}.
The paper is structured in the following way:
First, related work is summarized in Sec.~\ref{sec:related-work}, followed by the foundations in Sec.~\ref{sec:foundations}, after which the identified problem is further analyzed in Sec.~\ref{sec:problem_formulation}.
Then, the proposed methods are explained in Sec.~\ref{sec:method} followed by the evaluation in Sec.~\ref{sec:eval}. Finally, the paper is concluded in Sec.~\ref{sec:conclusion}.

%% file: plots/introduction/overview.tex
\pgfplotstableread[col sep=comma,]{plots/introduction/overview_driven.csv}\driventable

\pgfplotstableread[col sep=comma,]{plots/introduction/overview_corridor.csv}\corridortable
\pgfplotstableread[col sep=comma,]{plots/introduction/overview_goal_pose.csv}\goaltable

\begin{tikzpicture}
    \begin{axis}[
        width=1.0\linewidth,
        height=7.5cm,
        axis line style={draw=none},
        tick style={draw=none},
        xticklabel=\empty,
        yticklabel=\empty,
        axis equal,
        clip marker paths=true,
        xmax=20,
        ymin=0,
        ymax=30,
        clip=false,
    ]


    \node (C) 
    [
        draw=matlabYellow, 
        rounded corners,
        line width=0.4mm,
        text width=2.5cm,
        align=center,
        anchor=east,
        ] at 
        (-5.5, 24.0) {\small1. Dynamic Weight Allocation};
    \node (D) [
        circle,
        draw=matlabYellow,
        line width=0.4mm,
        minimum size=45pt,
        anchor=center,
    ] at (0.0, 24.0) {};

    \node (E) 
    [
        draw=matlabGreen, 
        rounded corners,
        line width=0.4mm,
        text width=3cm,
        align=center,
        anchor=north,
        ] at 
        (11.0, 1.8) {\small2. Dynamic Objective Allocation};
    \node (F) [
        circle,
        draw=matlabGreen,
        line width=0.4mm,
        minimum size=45pt,
        anchor=center,
    ] at (11.0, 6.0) {};

    \addplot[
    color=black, 
    line width=1pt,
    color=black,
    each nth point=5,
    ] 
    table [
        x=x_bound_left, 
        y=y_bound_left, 
    ] {\corridortable};
    
    \addplot[
    color=black, 
    line width=1pt,
    color=black,
    each nth point=5,
    ] 
    table [
        x=x_bound_left, 
        y=y_bound_left, 
    ] {\corridortable};
    
    \addplot[
        color=black, 
        line width=1pt,
        color=black, 
        each nth point=5,
        forget plot
    ] 
    table [
        x=x_bound_right, 
        y=y_bound_right, 
    ] {\corridortable};

    \addplot[
    color=newGray, 
    line width=1pt,
    each nth point=5,
    ] 
    table [
        x=path_x, 
        y=path_y, 
    ] {\corridortable};

    \addplot[
        color = matlabBlue,
    line width=1.0mm,
        each nth point = 3,
    ] 
    table [
        x=sim_x, 
        y=sim_y,
    ] {\driventable};
    
    \addplot[
        line width=1.5pt,
        color=matlabDarkRed
    ] 
    table [x=x, y=y] {\goaltable};
    
    \addplot[
        line width=1.5pt,
        color=matlabDarkRed
    ] 
    table [x=rl_x, y=rl_y] {\goaltable};
    \addplot[
        line width=1.5pt,
        color=matlabDarkRed
    ] 
    table [x=rr_x, y=rr_y] {\goaltable};
    \addplot[
        line width=1.5pt,
        color=matlabDarkRed
    ] 
    table [x=fl_x, y=fl_y] {\goaltable};
    \addplot[
        line width=1.5pt,
        color=matlabDarkRed
    ] 
    table [x=fr_x, y=fr_y] {\goaltable};
    \end{axis}
\end{tikzpicture}

%% file: doc/02-related-work.tex
\FloatBarrier
\section{Related Work}\label{sec:related-work}
The first topic that is handled in this work is the problem of path-following. 
Hence, we assume that a global kinematically feasible path has already been planned.
Possible methods to follow this path are the Stanley controller~\cite{thrun2006stanley} and pure pursuit~\cite{coulter1992pure} or regulated pure pursuit~\cite{macenski2023regulated} controller.
These provide a control law that stabilizes the robot along the path.
Other approaches are sampling-based, like the dynamic window approach (DWA)~\cite{fox1997dwa} or model predictive path integral (MPPI)~\cite{williams2016mppi} method.
At last, model predictive control (MPC)~\cite{ziegler2014bertha} and especially model predictive contouring control (MPCC)~\cite{liniger2015, romero2022} can be used in which the path following problem is solved using numerical optimization. 
Especially, \cite{romero2022} proposes a method to dynamically change the weights of an MPC-based controller, which is used to fly a drone precisely through a number of goals.

The next topic is the task of navigating to a specific pose.
At first, this goal can be achieved by planning a path to this pose followed by a path-following controller.
However, if the goal pose is derived from noisy measurements, this method is not suitable as it would trigger a frequent replanning of a new path, which is resource-intensive and could lead to varying paths caused by discretization errors in the path planning algorithm, which reduces the robustness.
Therefore, methods to plan directly to a goal pose are necessary. 
They are the graceful controller~\cite{park2011graceful}, which is unable to handle non-holonomic system dynamics, flatness-based controllers~\cite{fuchshumer2005flatness}, (iterative) linear quadratic regulators~\cite{chen2019ilqr}, and also MPC-based methods~\cite{zhang2018opt}.
The latter ones are able to handle non-holonomic system dynamics as present in normal vehicles.

However, there are scenarios in which both methods should be applied shortly after each other. 
This is the case, e.g., in robot docking maneuvers to their charging stations or if they must navigate to pick up a package.
Usually, the path-following step is completely separated from the step to plan to a pose, like in the docking server of the Nav2 stack~\cite{macenski2020marathon2, macenski2023survey}, which is used in the industry by different robot manufacturers like Dexory~\cite{dexory}.
The Nav2 stack uses the graceful controller mentioned above, which has the disadvantage of being incapable of handling non-holonomic vehicle dynamics.
Due to this limitation and the possibility of first navigating to a staging pose, this planning framework can generate suboptimal trajectories.
Hence, it would be advantageous to address this topic with a unified approach.
To the best of our knowledge, no motion planning algorithm yet exists that solves the two problems of path following and moving to a pose in one go.
As stated above, MPC-based methods can solve both problems.
Therefore, this paper proposes an approach using a variant of MPC that can solve both problems inherently.

%% file: doc/03-foundations.tex
\section{Foundations}
\label{sec:foundations}
This section gives a brief overview of the concept of MPC and MPCC.
The fundamental algorithm in this work is an MPCC similar to the ones in~\cite{liniger2015} and~\cite{romero2022}. 
Hence, the reader is referred to~\cite{romero2022} for further details.

The core of MPC is an optimal control problem (OCP).
This section will define the terms of an OCP partly discretely and continuously to allow easier comparison with its implementation.
The OCP is solved for a certain time horizon $T$, which is divided into $N$ steps. 
The sub-index $k\in N$ specifies variables at a single discretized time step.
Its goal is to find a sequence of inputs $\boldsymbol{u}_k$ that minimizes a given cost function $J(\boldsymbol{x})$ depending on the state and input vectors~$\boldsymbol{x} \in \mathcal{X}$ and~$\boldsymbol{u} \in \mathcal{U}$ with respect to the dynamic function of a given system $f: \mathcal{X} \times \mathcal{U} \mapsto \mathcal{X}$.
\begin{align}
\begin{split}
    \argminA_u\quad &J(\boldsymbol{x}) \\
    \text{subject to} \quad &\boldsymbol{x}_0 = \boldsymbol{x} \\&\boldsymbol{x}_{k+1} = f(\boldsymbol{x}_k, \boldsymbol{u}_k) \\
    \end{split}
\end{align}
The standard cost function minimizes the quadratic differences of the states and inputs to a given reference $\boldsymbol{x}_k^\text{r}$, $\boldsymbol{u}_k^\text{r}$.
These differences are denoted by $\Delta\boldsymbol{x}_k$, $\Delta\boldsymbol{u}_k$ and $\Delta\boldsymbol{x}_N$. Thus, the cost function is defined by
\begin{align}
    J(\boldsymbol{x}) = 
        \Vert \Delta\boldsymbol{x}_N \Vert^2_{\boldsymbol{Q}_N} +  
        \sum_{k=0}^{N-1}
        \Vert \Delta\boldsymbol{x}_k \Vert^2_{\boldsymbol{Q}} + 
        \Vert \Delta\boldsymbol{u}_k \Vert^2_{\boldsymbol{R}}\,. \label{eq:mpc}
\end{align}
The matrices $\boldsymbol{Q}_N=\diag(\boldsymbol{q_N})$, $\boldsymbol{Q}=\diag(\boldsymbol{q})$, $\boldsymbol{R}=\diag(\boldsymbol{r})$ are the weight matrices for the weighted Euclidian product denoted by $\Vert \boldsymbol{x}\Vert^2_Q = \boldsymbol{x}^T\cdot \boldsymbol{Q} \cdot \boldsymbol{x}$.
This summarizes the structure of a standard MPC.

The MPCC approach comprises some additional changes.
The state vector $\boldsymbol{x}$ is extended by an additional state $\theta$ that specifies the arc length along a given reference path that should be followed.
This reference path is a parameterized path denoted by $\boldsymbol{p}^\text{r}(\theta) = (x^\text{r}(\theta), y^\text{r}(\theta), \phi^\text{r}(\theta))$.
In addition, the input vector $\boldsymbol{u}$ is extended by the derivative of the arc length $\dot\theta$.
Hence, each state has a corresponding reference pose $\boldsymbol{p}_k^r=(x_k^\text{r}, y_k^\text{r}, \phi_k^\text{r})$ whose location can be controlled by the virtual input $\dot\theta$.
This input is penalized with the parameter $\gamma$, which is a negative penalty, hence a reward, that causes the MPCC to maximize the value of $\dot\theta$, leading to a progression of the reference poses along the reference path.
The particularity of the MPCC is that it contains a vector of additional Frenet coordinates $\boldsymbol{x}^\text{F}$ with their corresponding weight matrix $\boldsymbol{Q}^\text{F}=\diag(\boldsymbol{q}^\text{F})$.
The frenet coordinates and the corresponding weight vector are defined by
\begin{align}
        \boldsymbol{x}^\text{F} = \begin{pmatrix} 
        e^\text{l} \\ 
        e^\text{c} \\ 
        \end{pmatrix}\,,\quad 
    \boldsymbol{q}^\text{F} = 
    \begin{pmatrix} 
    q^\text{l} \\
    q^\text{c} \\ 
    \end{pmatrix} \,. \label{eq:frenet_state_and_weight}
\end{align}
The Frenet coordinates can be approximated by the following equation, which denotes a rotation of the coordinate differences around the reference pose.
\begin{align}
        \begin{pmatrix} 
        e^\text{l} \\ 
        e^\text{c} \\ 
        \end{pmatrix}=
        \begin{pmatrix}
        (x\!-\!x^\text{r})\!\cdot\!\cos(\phi^r)\!+\!(y\!-\!y^\text{r})\!\cdot\!\sin(\phi^\text{r})\\ 
        -(x\!-\!x^\text{r})\!\cdot\!\sin(\phi^r)\!+\!(y\!-\!y^\text{r})\!\cdot\!\cos(\phi^\text{r})\\ 
        \end{pmatrix}\,.
        \label{eq:contouring}
\end{align}
However, this approximation is only valid if $\boldsymbol{x}^\text{F}$ if $q^\text{l} \gg q^\text{c}$.
Only in this case is $e^\text{c}$ equal to the lateral deviation to the reference pose $\boldsymbol{p^\text{r}}$ and $e^\text{l}$ to the longitudinal deviation.
Hence, the MPCC maximizes its progress $\dot\theta$ while minimizing the longitudinal deviation to the reference pose, which provides a valid lateral distance to the reference path encoded in the Frenet coordinate $e^\text{c}$.
Usually, there is no final position to reach.
Hence, the term $\Vert\Delta\boldsymbol{x}_N\Vert^2_{\boldsymbol{Q}_N}$ is often omitted as no reference for the final state exists.
The overall cost function of the MPCC is then denoted by
\begin{align}
\begin{split}
    J(\boldsymbol{x}) = 
        \sum_{k=0}^{N-1}
        &\Vert \Delta\boldsymbol{x}_k \Vert^2_{\boldsymbol{Q}} + 
        \Vert \Delta\boldsymbol{x}^\text{F}_k \Vert^2_{\boldsymbol{Q^\text{F}}} + \\
        &\Vert \Delta\boldsymbol{u}_k \Vert^2_{\boldsymbol{R}} +
        \dot\theta_k \gamma \,. \label{eq:mpcc}
\end{split}
\end{align}

%% file: doc/04-problem-formulation.tex
\section{Problem Formulation}\label{sec:problem_formulation}
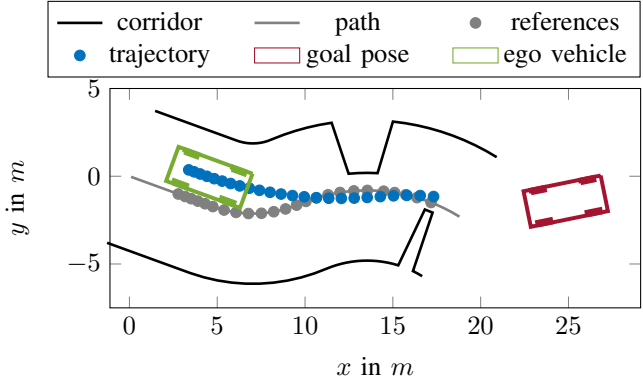
\begin{figure}[t]
    \centering
    \vspace{2mm}
    \input{plots/problem/problem}%
    \caption{%
    Visualization of the stated problem.
    Each trajectory state $\boldsymbol{x}_k$ has a reference pose $p_k^\text{r}$ perfectly orthogonally at the reference path.
    This provides valid frenet coordinates.
    However, with standard MPCC, the goal pose behind the corridor cannot be reached without switching to another motion planning algorithm, as there is no valid pairing of a state and a reference pose for $\theta > \theta^\text{e}$. 
    }%
    \label{fig:problem}
\end{figure}%
The automated vehicle controlled in this work can be approximated by a bicycle mode. 
Thus, the state vector $\boldsymbol{x}$ and the input vector $\boldsymbol{u}$, including their extensions by MPCC, are defined by
\begin{align}
    \boldsymbol{x} = \begin{pmatrix} 
    x\\ y\\ \phi \\ v\\ \delta \\ \theta \\ 
    \end{pmatrix}, 
        \quad
    \boldsymbol{u} = \begin{pmatrix} 
    a\\ 
    \dot{\delta} \\
    \dot{\theta} %
    \end{pmatrix}\,.  
\end{align}
Further, the problem solved in this work is visualized in Fig.~\ref{fig:problem}.
The goal of the trajectory planning algorithm is to follow a path under certain state and input constraints. 
They are, among others, the so-called corridor constraints. 
They define the extent of obstacles in Frenet coordinates and constrain the contouring error by $\underline{e}^\text{c}(\theta) < e^\text{c}(\theta) < \overline{e}^\text{c}(\theta)$.
However, goal poses extracted from noisy measurements change over time, leading to goal poses that are different from the pose that the underlying path was previously planned to.
The motion planning algorithm should be able to plan to these poses without having to repeat the path planning approach.
This should also be possible for poses that lie behind the path end denoted by $\theta^\text{e}$. 
Thus, the trajectory must be able to have states with $\theta > \theta^\text{e}$.

%% file: plots/problem/problem.tex
\pgfplotstableread[col sep=comma,]{plots/problem/problem_trajectory.csv}\trajectorytable

\pgfplotstableread[col sep=comma,]{plots/problem/problem_corridor.csv}\corridortable
\pgfplotstableread[col sep=comma,]{plots/problem/problem_goal_pose.csv}\goaltable
\pgfplotstableread[col sep=comma,]{plots/problem/problem_ego_pose.csv}\egotable

\begin{tikzpicture}
    \begin{axis}[
        width=1.0\linewidth,
        height=4.5cm,
        xlabel=$x$ in $m$,
        ylabel=$y$ in $m$,
        axis equal,
        clip marker paths=true,
        xmin=0,
        xmax=28,
        ymin=-7.5,
        ymax=5,
        clip=false,
        legend columns = 3,
        legend style={at={(1.0, 1.05)},anchor=south east, /tikz/every even column/.append style={column sep=0.5cm}},
    ]
    
    \addplot[
    color=black, 
    line width=1pt,
    color=black,
    each nth point=5,
    ] 
    table [
        x=x_bound_left, 
        y=y_bound_left, 
    ] {\corridortable};
    \addlegendentry{corridor}
    
    \addplot[
        color=black, 
        line width=1pt,
        color=black, 
        each nth point=5,
        forget plot
    ] 
    table [
        x=x_bound_right, 
        y=y_bound_right, 
    ] {\corridortable};

    \addplot[
    color=newGray, 
    line width=1pt,
    each nth point=5,
    ] 
    table [
        x=path_x, 
        y=path_y, 
    ] {\corridortable};
    \addlegendentry{path}

    \addplot[
        only marks,
        color = gray,
        each nth point = 3,
    ] 
    table [
        x=xr, 
        y=yr,
    ] {\trajectorytable};
    \addlegendentry{references}
    
    \addplot[
        only marks,
        color = matlabBlue,
        each nth point = 3,
    ] 
    table [
        x=traj_x, 
        y=traj_y,
    ] {\trajectorytable};
    \addlegendentry{trajectory}
    
    \addplot[
        line width=1.5pt,
        color=matlabDarkRed,
        forget plot,
    ] 
    table [x=x, y=y] {\goaltable};
    \addlegendimage{area legend, matlabDarkRed}
    \addlegendentry{goal pose}
    
    \addplot[
        line width=1.5pt,
        color=matlabDarkRed,
        forget plot
    ] 
    table [x=rl_x, y=rl_y] {\goaltable};
    \addplot[
        line width=1.5pt,
        color=matlabDarkRed,
        forget plot
    ] 
    table [x=rr_x, y=rr_y] {\goaltable};
    \addplot[
        line width=1.5pt,
        color=matlabDarkRed,
        forget plot
    ] 
    table [x=fl_x, y=fl_y] {\goaltable};
    \addplot[
        line width=1.5pt,
        color=matlabDarkRed,
        forget plot
    ] 
    table [x=fr_x, y=fr_y] {\goaltable};

    \addplot[
        line width=1.5pt,
        color=matlabGreen,
        forget plot
    ] 
    table [x=x, y=y] {\egotable};
    \addlegendimage{area legend, matlabGreen}
    \addlegendentry{ego vehicle}
    
    \addplot[
        line width=1.5pt,
        color=matlabGreen
    ] 
    table [x=rl_x, y=rl_y] {\egotable};
    \addplot[
        line width=1.5pt,
        color=matlabGreen
    ] 
    table [x=rr_x, y=rr_y] {\egotable};
    \addplot[
        line width=1.5pt,
        color=matlabGreen
    ] 
    table [x=fl_x, y=fl_y] {\egotable};
    \addplot[
        line width=1.5pt,
        color=matlabGreen
    ] 
    table [x=fr_x, y=fr_y] {\egotable};
    \end{axis}
\end{tikzpicture}

%% file: doc/05-method.tex
\FloatBarrier
\section{Method}\label{sec:method}
This section explains the methods of our contributions.

\subsection{Dynamic Weight Allocation}\label{sec:method:dyn_weight}
In path-following problems, the path ends must be reached precisely if they are the goal of the overall motion plan. 
Further, the cusp points, at which direction switches are necessary, should also be reached precisely to ensure that the path can be followed with respect to the vehicle's system dynamics.
Therefore, we introduce a dynamic weight allocation technique similar to the one used by~\cite{romero2022} to precisely reach the mentioned points.

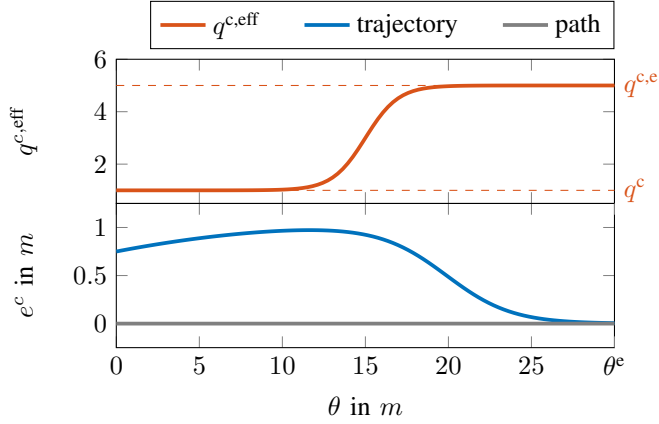
\begin{figure}[t]
    \centering
    \vspace{2mm}
    \input{plots/method/dyn_weight/dyn_weight}%
    \caption{%
    Schematic visualization of the dynamic weight allocation method to precisely reach the path end at $\theta^\text{e}$: The sigmoid function $\sigma(\theta)$ blends the weights $q^\text{c}$ into $q^\text{c,e}$ for trajectory states approaching the path end at $\theta=30$.
    This causes an increasing penalty of $e^\text{c}$, which leads to the blue trajectory in Frenet coordinates with minimized $e^\text{c}$ towards the path end.
    }%
    \label{fig:dyn_weight}
\end{figure}

\begin{figure}[t]
    \centering
    \vspace{2mm}
    \input{plots/method/dyn_weight_goal/dyn_weight_goal}%
    \caption{%
    Schematic visualization of the dynamic weight allocation to reach a goal pose inside the corridor: 
    The weights $\gamma^\text{eff}$ and $q^\text{c,eff}$ of states close to the projected longitudinal goal position $\theta^\text{g}$ are blended to zero.
    Now, the penalty $\boldsymbol{q}_N^\text{eff}$ to reach the goal pose dominates the behavior of the MPC.
    Hence, the algorithm plans precisely to the goal pose while being able to deviate from the reference path which is shown by the blue trajectory in Cartesian coordinates.
    }%
    \label{fig:method:dyn_weight_goal}
\end{figure}
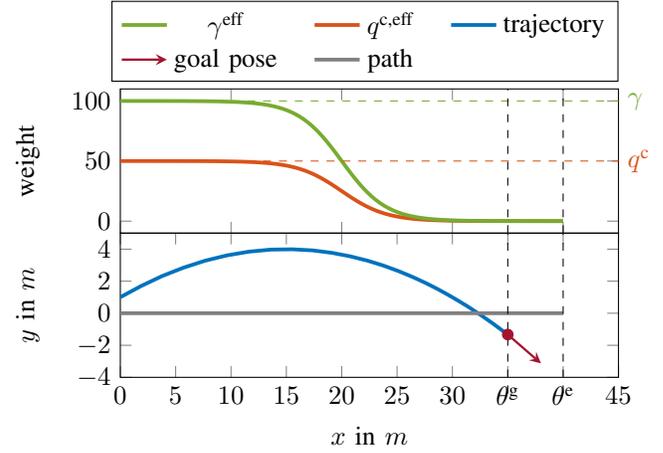

The first part of the method is setting weight $q^\text{c}$ for the contouring error $e^\text{c}$ to the path depending on the proximity to the end of the path. 
This distance is calculated by the longitudinal position of the trajectory given by $\theta$ to the longitudinal position of the path end or switching point, which is denoted by $\theta^\text{e}$.
The resulting distance is denoted by $\epsilon(\theta)$.
Now an effective $q^\text{c,eff}$ is calculated by blending the original contouring weight $q^\text{c}$ with an increased one, denoted by $q^\text{c,e}$, by means of a sigmoid function $\sigma(\theta)$:
\begin{align}
\begin{split}
    \epsilon(\theta) &= \theta^\text{e} - \theta \\
    \sigma(\theta) &= \frac{1}{1+e^{\,\alpha\cdot(\epsilon(\theta) - \beta)}} \\
    q^\text{c,eff}(\theta) &= \sigma(\theta) \cdot q^\text{c,e} + (1-\sigma(\theta)) \cdot q^\text{c}.\label{eq:qceff}
    \end{split}
\end{align}
Here, $\alpha$ and $\beta$ serve as tuning parameters to set the steepness and the offset of the sigmoid function, which defines the proximity in which the cost starts to blend, leading to the vehicle staying closer to the reference path.
Hence, the effective contouring weight vector is changed from the fundamental version in Eq.~\ref{eq:frenet_state_and_weight} to
\begin{align}
    \boldsymbol{q}^\text{F}(\theta) = 
    \begin{pmatrix} 
    q^\text{l} \\
    q^\text{c,eff}(\theta) \\ 
    \end{pmatrix}\,.
\end{align}
Fig.~\ref{fig:dyn_weight} visualizes the key aspects of this method schematically.
The figure visualizes the rise of $q^\text{c,eff}$ by the dynamic weight allocation for trajectory states near the path end $\theta^e$.
This causes the trajectory to minimize the contouring error $e^\text{c}$ towards the path end.
This method can now be applied to another use case.
If the goal of the motion plan is a pose $\boldsymbol{p}^\text{g}$ that lies not on the path but in its proximity, the weights can be adapted similarly. 
At first, the goal pose must be projected to the path, which yields its longitudinal position~$\theta^\text{g}$.
The weight $q^\text{c}$, which was increased before, must now be deactivated on approach to allow the vehicle to deviate from the path.
Further, also the negative weight of the reward $\gamma$ to increase the progress along the reference path must be blended out because otherwise, the MPC would try to reach the path end. 
Hence, also an effective $\gamma^\text{eff}$ is calculated.
This dynamic weight allocation is then calculated similarly as in Eq.~\eqref{eq:qceff} by
\begin{align}
\begin{split}
    \epsilon(\theta) &= \theta^\text{g} - \theta \\
    q^\text{c,eff}(\theta) &= (1-\sigma(\theta)) \cdot q^\text{c} \\
    \gamma^\text{eff}(\theta) &= (1-\sigma(\theta)) \cdot \gamma \, .\label{eq:qceff_inside_goal}
    \end{split}
\end{align}
In addition to that, we now reinsert the cost term belonging of the last trajectory state $\Delta\boldsymbol{x}_N$ to reach a specific reference position $\boldsymbol{x}_N^\text{r}$ which is set by the mentioned goal pose $\boldsymbol{p}^\text{g}$.
Hence, the weight vector $\boldsymbol{q}_\text{N}$ must be changed as well into an effective $\boldsymbol{q}_N^\text{eff}$.
Here, this weight is blended similarly as before, but in this case, it depends on the longitudinal position of the goal along the path $\theta^\text{g}$ to the longitudinal position of the base of the robot, which is denoted by $\theta_0$, and not of the trajectory states.
An equal approach is not possible, as increasing the costs to reach the final goal causes an intermediate rise in the overall costs, which forces the trajectory to avoid the proximity of the goal completely.
Consequently, the weights to reach the goal are calculated by
\begin{align}
\begin{split}
    \epsilon_0 &= \theta^\text{e} - \theta_0 \\
    \sigma &= \frac{1}{1+e^{\,\alpha\cdot(\epsilon_0 - \beta)}} \\
    \boldsymbol{q}_N^{\text{eff}} &= \sigma \cdot \boldsymbol{q}_N^\text{e} + (1-\sigma)\cdot \boldsymbol{q}_N \, .
        \end{split}
\end{align}
With this formulation, the MPCC is able to reach goal poses that are longitudinally within the corridor.
The dynamic weight allocation approach to reach the goal pose is also visualized schematically in Fig.~\ref{fig:method:dyn_weight_goal}.

\subsection{Dynamic Objective Allocation}\label{sec:method:dyn_cost_type}
The method just proposed cannot be applied directly to scenarios in which the goal pose $\boldsymbol{p}^\text{g}$ lies longitudinally behind the corridor.
In these cases, the MPCC cannot reach the pose, as the high penalty by $q^\text{l}$ prevents it from leaving the corridor longitudinally.
Hence, we apply the dynamic weight allocation method to the weight of the longitudinal error $q^\text{l}$ and set it to zero for trajectory states that lie behind the path end ($\theta > \theta^\text{e}$).
If the states lie within the corridor, we keep the default value $q^\text{l}$ to guarantee a valid Frenet state $\boldsymbol{x}^\text{F}$ at each stage.
This leads to the following behavior: 
Behind the corridor, the contouring aspect of the MPCC is deactivated, and only the Cartesian penalty $\boldsymbol{q}_N$ of the final state $\boldsymbol{x}_N$ determines the behavior of the MPCC, which tries to reach the goal pose.
This case distinction can be defined by
\begin{align}
    q^\text{l,eff} = \begin{cases} q^\text{l} \quad &\text{for} \quad \theta < \theta^\text{e} \\
    0 \quad &\text{for} \quad \theta \ge \theta^\text{e}\end{cases} \,. \label{eq:case_dist}
\end{align}
Further, all possible penalties or constraints related to Frenet coordinates must be deactivated as well because there are no valid Frenet coordinates behind the path end.
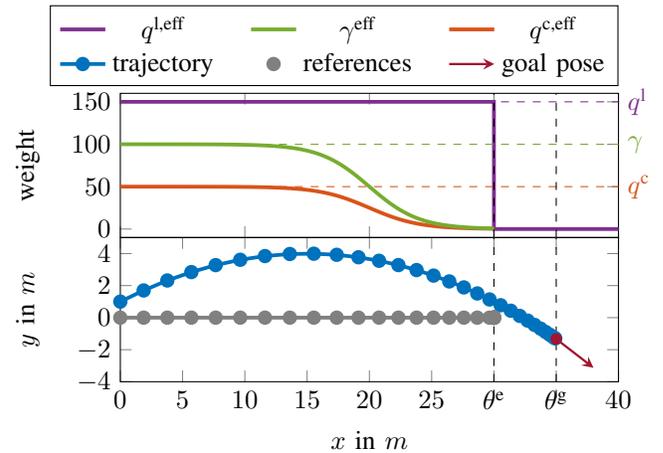
\begin{figure}[b]
    \centering
    \vspace{2mm}
    \input{plots/method/dyn_cost_type/dyn_cost_type}%
    \caption{%
    Schematic visualization of the dynamic objective allocation: The purple curve at the top visualizes the drop of $q^\text{l, eff}$ caused by the case distinction from Eq.~\eqref{eq:case_dist}. 
    The other weights are equally blended as in Fig.\ref{fig:method:dyn_weight_goal}
    In the lower part, the trajectory of a vehicle maneuvering to a goal pose is shown. 
    For every state $\boldsymbol{x}$ with a $\theta \le \theta^\text{e}$, a corresponding reference exists on the reference path. 
    For states behind the corridor with $\theta < \theta^\text{e}$, $q^\text{l, eff}$ is set to zero, which allows them to have no valid reference.
    }%
    \label{fig:dyn_cost_type}
    \vspace{-0.2cm}
\end{figure}%
To summarize, a valid reference exists only for states inside the corridor.
But for $\theta > \theta^e$, this pairing is no longer enforced, and the states are penalized as if they were part of a pure Cartesian MPC.
Because this dynamic weight allocation not only increases or decreases the magnitude of weights but also completely changes the main objective of a trajectory state, we call this method dynamic objective allocation.
The complete approach is visualized schematically in Fig.~\ref{fig:dyn_cost_type}.

%% file: plots/method/dyn_weight/dyn_weight.tex


\def\q{1}
\def\qe{5}
\def\thetamax{30}
\def\ymax{\qe+1}

\def\A{-1}
\def\B{15}
\def\sigmoid{1/(1.0 + exp(\A * (x - \B)))}

\def\modifier{-1/(15*15*4) * (x-15)^2 + 1}
\def\sigmoidtraj{1/(1.0 + exp(0.5 * (x - 20)))}
\def\trajectory{\sigmoidtraj * (\modifier)}

\begin{tikzpicture}
    \begin{groupplot}[
        group style={
            group size=1 by 2,
            x descriptions at=edge bottom,
            vertical sep=0pt,
        },
    ]
   
    \nextgroupplot[
    xmin=0,
    xmax=\thetamax,
    xmin=0,
    ymax=\ymax,
    width=0.95\linewidth,
    height=3.5cm,
    ylabel=$q^{c,\text{eff}}$,
    clip = false,
    ]
        \addplot[
        color=matlabRed, 
        line width=0.5mm,
        domain=0:\thetamax,
        samples=100,
        ]{\sigmoid * \qe + (1-\sigmoid) * \q};
        
        \draw[color=matlabRed, dashed] 
            (axis cs:0, \q) -- (axis cs:\thetamax, \q);
        \draw[color=matlabRed, dashed] 
            (axis cs:0, \qe) -- (axis cs:\thetamax, \qe);

        \node [text=matlabRed, anchor=west] at (axis cs:\thetamax, \q) {$q^\text{c}$};
        \node [text=matlabRed, anchor=west] at (axis cs:\thetamax, \qe) {$q^\text{c,e}$};
    
    \nextgroupplot[
        xmin=0,
        xmax=\thetamax,
        width=0.95\linewidth,
        height=3.5cm,
        ymin=-0.25,
        ymax=1.25,
        xlabel=$\theta$ in $m$,
        ylabel=$e^c$ in $m$,
        clip = false,
        xtick={0, 5, 10, 15, 20, 25},
        extra x ticks={\thetamax},
        extra x tick labels={$\theta^\text{e}$},
        legend columns=3,
        legend style={at={(1.0, 2.07)},anchor=south east, /tikz/every even column/.append style={column sep=0.5cm}},
    ]

        \addlegendentry{$q^\text{c,eff}$}
        \addlegendimage{no markers, line width=0.5mm, matlabRed}
            
        \addplot [
        color=matlabBlue, 
        line width=0.5mm,
        domain=0:\thetamax,
        samples=100,
        ]{\trajectory};
        \addlegendentry{trajectory}

       \addplot[
        color=newGray, 
        line width=0.5mm,
        ] coordinates {
            (0,0) (\thetamax,0)
        }; 
        ] coordinates {
            (0,0) (\thetamax,0)
        }; 
        \addlegendentry{path}
        
    \end{groupplot}
\end{tikzpicture}%

%% file: plots/method/dyn_weight_goal/dyn_weight_goal.tex

\pgfplotstableread[col sep=comma,]{plots/method/dyn_cost_type/schematic_goal_pose.csv}\schematicgoalpose

\def\thetamax{40}
\def\thetagoal{35}

\def\qc{50}
\def\qce{0}
\def\A{-0.5}
\def\B{20}
\def\sigmoid{1/(1.0 + exp(\A * (x - \B)))}

\def\gammastd{100}
\def\gammae{0}

\def\xmax{\thetamax + 5}
\def\ymin{-10}
\def\ymax{\gammastd*1.1}

\begin{tikzpicture}
    \begin{groupplot}[
        group style={
            group size=1 by 2,
            x descriptions at=edge bottom,
            vertical sep=0pt,
        },
    ]
   
    \nextgroupplot[
    xmin=0,
    xmax=\xmax,
    xmin=0,
    ymin=\ymin,
    ymax=\ymax,
    width=0.95\linewidth,
    height=3.5cm,
    ylabel=weight,
    clip = false,
    ]

        \addplot[
            color=matlabRed, 
            line width=0.5mm,
            domain=0:\thetamax,
            samples=100,
        ]{\sigmoid * \qce + (1-\sigmoid) * \qc};

        \addplot[
            color=matlabGreen, 
            line width=0.5mm,
            domain=0:\thetamax,
            samples=100,
        ]{\sigmoid * \gammae + (1-\sigmoid) * \gammastd};
        
        \draw[color=matlabRed, dashed] 
            (axis cs:0, \qc) -- (axis cs:\xmax, \qc);
        \draw[color=matlabGreen, dashed] 
            (axis cs:0, \gammastd) -- (axis cs:\xmax, \gammastd);

        \draw[color=black, dashed] 
            (axis cs:\thetamax, \ymin) -- (axis cs:\thetamax, \ymax);
        \draw[color=black, dashed] 
            (axis cs:\thetagoal, \ymin) -- (axis cs:\thetagoal, \ymax);

        \node [text=matlabRed, anchor=west] at (axis cs:\xmax, \qc) {$q^\text{c}$};
        \node [text=matlabGreen, anchor=west] at (axis cs:\xmax, \gammastd) {$\gamma$};
    
    \nextgroupplot[
        xmin=0,
        xmax=\xmax,
        width=0.95\linewidth,
        height=3.5cm,
        ymin=-4,
        ymax=5,
        xlabel=$x$ in $m$,
        ylabel=$y$ in $m$,
        clip = false,
        xtick={0, 5, 10, 15, 20, 25, 30, 45},
        extra x ticks={\thetagoal, \thetamax},
        extra x tick labels={$\theta^\text{g}$, $\theta^\text{e}$},
        legend columns=3,
        legend style={at={(1.0, 2.05)},anchor=south east, /tikz/every even column/.append style={column sep=0.4cm}},
    ]

        \addlegendentry{$\gamma^\text{eff}$}
        \addlegendimage{no markers, line width=0.5mm, matlabGreen}
        
        \addlegendentry{$q^\text{c,eff}$}
        \addlegendimage{no markers, line width=0.5mm, matlabRed}
        
        \addplot [
        matlabBlue,
        line width=0.5mm,
        ]table [
            x=x, 
            y=y,
        ] {\schematicgoalpose};
        \addlegendentry{trajectory}

        \addplot [
        quiver={
            u=1,v=-0.6, 
            scale arrows=3.0,
            every arrow/.append style={line width=0.3mm},
        },
        matlabDarkRed, 
        -stealth,
        ] coordinates {
            (\thetagoal, -1.33)
        };
        \addlegendentry{goal pose}

        \addplot[
        color=newGray, 
        line width=0.5mm,
        ] coordinates {
            (0,0) (\thetamax,0)
        }; 
        \addlegendentry{path}

        \addplot[
        only marks,
        matlabDarkRed,
        mark=*, 
        forget plot,
        ] coordinates {
            (\thetagoal, -1.33)
        };
        
        \draw[color=black, dashed] 
            (axis cs:\thetamax, -5) -- (axis cs:\thetamax, 5);
        \draw[color=black, dashed] 
            (axis cs:\thetagoal, -5) -- (axis cs:\thetagoal, 5);
    \end{groupplot}
\end{tikzpicture}%

%% file: plots/method/dyn_cost_type/dyn_cost_type.tex

\pgfplotstableread[col sep=comma,]{plots/method/dyn_cost_type/schematic_goal_pose.csv}\schematicgoalpose

\def\ql{150}
\def\thetamax{30}
\def\thetagoal{35}
\def\xmax{\thetagoal + 5}
\def\ymin{-10}
\def\ymax{\ql+10}

\def\qc{50}
\def\qce{0}
\def\A{-0.5}
\def\B{20}
\def\sigmoid{1/(1.0 + exp(\A * (x - \B)))}

\def\gammastd{100}
\def\gammae{0}


\begin{tikzpicture}
    \begin{groupplot}[
        group style={
            group size=1 by 2,
            x descriptions at=edge bottom,
            vertical sep=0pt,
        },
    ]
   
    \nextgroupplot[
    xmin=0,
    xmax=\xmax,
    xmin=0,
    ymin=\ymin,
    ymax=\ymax,
    width=0.95\linewidth,
    height=3.5cm,
    ylabel=weight,
    clip = false,
    ]
        \addplot[
        color=matlabPurple, 
        line width=0.5mm,
        ] coordinates {
            (0,\ql) (\thetamax,\ql) (\thetamax,0) (\xmax, 0)
        }; 

        \addplot[
            color=matlabRed, 
            line width=0.5mm,
            domain=0:\thetamax,
            samples=100,
        ]{\sigmoid * \qce + (1-\sigmoid) * \qc};

        \addplot[
            color=matlabGreen, 
            line width=0.5mm,
            domain=0:\thetamax,
            samples=100,
        ]{\sigmoid * \gammae + (1-\sigmoid) * \gammastd};

        \draw[color=matlabPurple, dashed] 
            (axis cs:0, \ql) -- (axis cs:\xmax, \ql);
        \draw[color=matlabRed, dashed] 
            (axis cs:0, \qc) -- (axis cs:\xmax, \qc);
        \draw[color=matlabGreen, dashed] 
            (axis cs:0, \gammastd) -- (axis cs:\xmax, \gammastd);

        \draw[color=black, dashed] 
            (axis cs:\thetamax, \ymin) -- (axis cs:\thetamax, \ymax);
        \draw[color=black, dashed] 
            (axis cs:\thetagoal, \ymin) -- (axis cs:\thetagoal, \ymax);

        \node [text=matlabPurple, anchor=west] at (axis cs:\xmax, \ql) {$q^\text{l}$};
        \node [text=matlabRed, anchor=west] at (axis cs:\xmax, \qc) {$q^\text{c}$};
        \node [text=matlabGreen, anchor=west] at (axis cs:\xmax, \gammastd) {$\gamma$};
        
    
    \nextgroupplot[
        xmin=0,
        xmax=\xmax,
        width=0.95\linewidth,
        height=3.5cm,
        ymin=-4,
        ymax=5,
        xlabel=$x$ in $m$,
        ylabel=$y$ in $m$,
        clip = false,
        xtick={0, 5, 10, 15, 20, 25, 40},
        extra x ticks={\thetagoal, \thetamax},
        extra x tick labels={$\theta^\text{g}$, $\theta^\text{e}$},
        legend columns=3,
        legend style={at={(1.0, 2.05)},anchor=south east, /tikz/every even column/.append style={column sep=0.4cm}},
    ]
        \addlegendentry{$q^\text{l,eff}$}
        \addlegendimage{no markers, line width=0.5mm, matlabPurple}

        \addlegendentry{$\gamma^\text{eff}$}
        \addlegendimage{no markers, line width=0.5mm, matlabGreen}
        
        \addlegendentry{$q^\text{c,eff}$}
        \addlegendimage{no markers, line width=0.5mm, matlabRed}
        
        \addplot [
        matlabBlue,
        mark=*,
        line width=0.5mm,
        ]table [
            x=x, 
            y=y,
        ] {\schematicgoalpose};
        \addlegendentry{trajectory}

        \addplot [
        newGray,
        only marks,
        mark=*,
        line width=0.5mm,
        ]table [
            x=theta, 
            y=thetas_y,
        ] {\schematicgoalpose};
        \addlegendentry{references}

        \addplot [
        quiver={
            u=1,v=-0.6, 
            scale arrows=3.0,
            every arrow/.append style={line width=0.3mm},
        },
        matlabDarkRed, 
        -stealth,
        ] coordinates {
            (\thetagoal, -1.33)
        };
        \addlegendentry{goal pose}

        \addplot[
        color=newGray, 
        line width=0.5mm,
        forget plot
        ] coordinates {
            (0,0) (\thetamax,0)
        }; 

        \addplot[
        only marks,
        matlabDarkRed,
        mark=*, 
        forget plot,
        ] coordinates {
            (\thetagoal, -1.33)
        };
        
        \draw[color=black, dashed] 
            (axis cs:\thetamax, -5) -- (axis cs:\thetamax, 5);
        \draw[color=black, dashed] 
            (axis cs:\thetagoal, -5) -- (axis cs:\thetagoal, 5);

    \end{groupplot}
\end{tikzpicture}%

%% file: doc/06-evaluation.tex
\FloatBarrier
\section{Evaluation}\label{sec:eval}
This section comprises the evaluation of the two proposed methods.
First, the evaluation of the dynamic weight allocation applied to path ends and direction switches, as well as goal poses, is shown, followed by the evaluation of the dynamic objective allocation approach.
All following evaluations are executed using the same parameters, which are shown in Table~\ref{tab:params}.
The MPC was created by using the Python API of acados \cite{acados2021}.
The runtimes are $\approx \qty{1.7}{\milli\second}$ for a standard MPC run, $\approx \qty{3}{\milli\second}$ if new references are inserted, which is done every \qty{100}{\milli\second}, and $\approx \qty{7}{\milli\second}$ if the MPC is reset and reinitialized completely.
These runtimes were determined on an AMD Ryzen 9 7950X with a base clock of \qty{4.5}{\giga\hertz}. 
\begin{table}[b]
	\caption{MPC Parameters}
    \vspace{-0.5cm}
	\label{tab:params}
	\begin{center}
		\begin{tabular}{c | c | c}
			\toprule
			Parameter & Description & Value \\
			\midrule
            $T$ & time horizon & \qty{7}{\second} \\
            $N$ & number of stages & 70 \\
            $\boldsymbol{q}$ & state weights & $[0, 0, 0, 0, 0, 0]$ \\
            $\boldsymbol{q}_N$ & goal pose weights & $[0, 0, 0, 0, 0, 0]$ \\
            $\boldsymbol{q}_N^\text{e}$ & goal pose weights at $\theta^\text{g}$ & $[1e4, 1e4, 1e4, 0, 0, 0]$ \\
            $\boldsymbol{r}$ & input weights & $[1e3, 100, 0]$ \\
            $\boldsymbol{q}^\text{F}$ & Frenet weights & $[1e3, 1.0, 0.0]$ \\
            $q^\text{c,e}$ & lateral weight at $\theta^\text{e}$ & $ 100$ \\
            $\gamma$ & penalty of progress $\dot\theta$ & -100 \\
            $\alpha$ & steepness of sigmoids & $1.0$ \\
            $\beta$ & center of sigmoids & \qty{10}{\meter} \\
			\bottomrule
		\end{tabular}
	\end{center}
\end{table}%



\subsection{Dynamic Weight Allocation}\label{sec:eval:dyn_weigth}
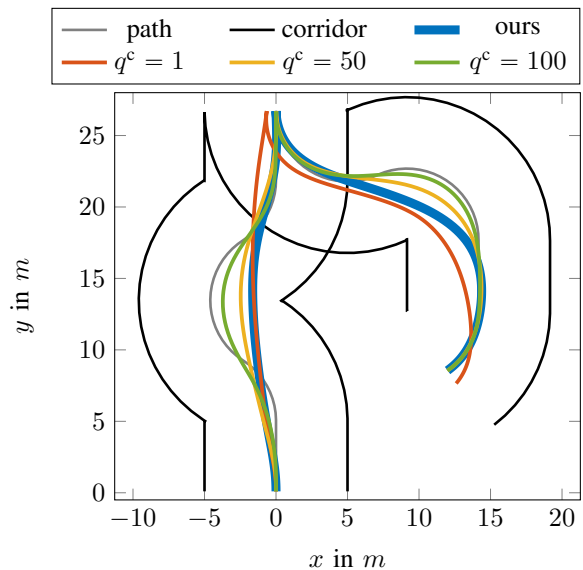
\begin{figure}[t]
    \centering
    \vspace{2mm}
    \input{plots/evaluation/dyn_weight/dyn_weight.tex}%
    \caption{%
    Driven trajectories for different constant lateral weights $q^\text{c}$ and our approach using the dynamic weight allocation method.
    Its effect can be observed when approaching the direction switch and the path end, which should be reached precisely.
    Our approach is able to reach both with high precision while smoothing out the other parts of the reference path.
    }%
    \label{fig:eval:dyn_weight}
\end{figure}%
In this section, the dynamic weight allocation is evaluated.
This is done with a path-following scenario that contains one direction switch.
At first, we compare our approach to the same MPC algorithm but with three different, fixed values of $q^\text{c}$.
This comparison is shown in Fig.~\ref{fig:eval:dyn_weight}.
It can be observed that with low values of $q^\text{c}$, the trajectory may vary from the reference path, leading to a smoother driven trajectory.
Usually, this is the intended behavior to generate a smooth and comfortable trajectory.
However, with low values of $q^\text{c}$, the trajectory does not reach the direction switches and the end of the path precisely.
In contrast to that, our approach can generate smooth trajectories along the reference path while reaching the direction changes and the path end precisely by adapting the weight $q^\text{c,eff}$ depending on the proximity of the state to these regions.
Hence, it inherently combines the advantages of low and high lateral penalties.

In addition to that, the dynamic weight allocation can be applied to reach goal poses inside a corridor. 
This objective is called Scenario~1 in the following.
The effect of our method is compared to two other methods. 
All are MPC-based methods that only differ in the approach used to reach the final pose. 

1. The MPCC plans to the path end or a certain distance in front of the goal while already using the dynamic weight allocation. 
After reaching the specified position, the MPCC is exchanged by a pure Cartesian MPC, which plans to the final pose. 
This method is called \emph{separated} in the following.
Its separated architecture is similar to the docking approach used in the Nav2 stack.

2. The MPCC plans along the corridor.
If the vehicle is close enough to the goal pose, a Cartesian MPC replaces the MPCC during runtime to reach it. 
This method is called \emph{switched} in the following.

\begin{figure}[t]
    \centering
    \vspace{2mm}
    \input{plots/evaluation/dyn_weight_goal/dyn_weight_goal}%
    \caption{%
    Scenario~1: Visualization of the second part of the corridor of Fig.~\ref{fig:eval:dyn_weight} after the switching point.
    Here, the goal pose lies within the corridor.
    \emph{Our} approach reaches this pose smoothly without excessive steering angles.
    The \emph{separated} baseline does not smooth the reference path and needs larger steering angles to reach the goal.
    Finally, the \emph{switched} baseline requires the vehicle to make an additional direction change.
    }%
    \label{fig:eval:dyn_weight_goal}
\end{figure}
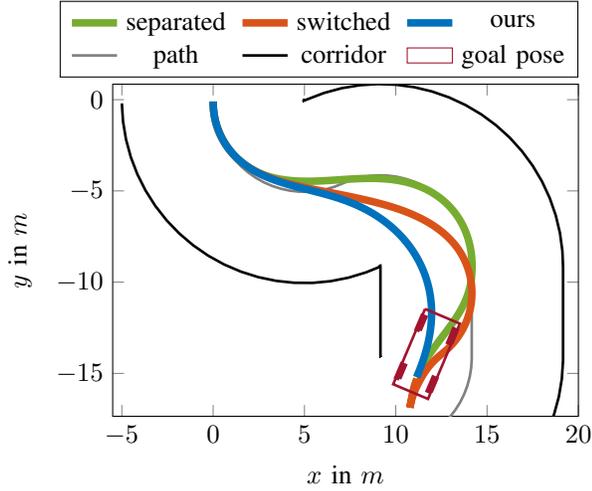

The driven trajectories of all three methods are visualized in Fig.~\ref{fig:eval:dyn_weight_goal}.
The \emph{separated} method reaches the goal without direction in contrast to the \emph{switched} approach, which does need an additional direction switch.
In contrast to that, our approach begins to plan to the goal pose early enough and reaches it smoothly.

\begin{figure}[t]
    \centering
    \vspace{2mm}
    \input{plots/evaluation/dyn_weight_goal/dyn_weight_velocity}%
    \caption{%
    Scenario 1: Driven velocities and steering angles of the scene in Fig.~\ref{fig:eval:dyn_weight_goal}, in which the goal pose lies within the corridor.
    \emph{Our} approach reaches the goal pose in about half the time compared to the \emph{separated} and \emph{switched} approaches while actuating lower steering angles $\delta$ and lower changes of the steering angle $\delta$, which can be estimated by the slope of the function.
    }%
    \label{fig:eval:dyn_weight_velocity}
\end{figure}
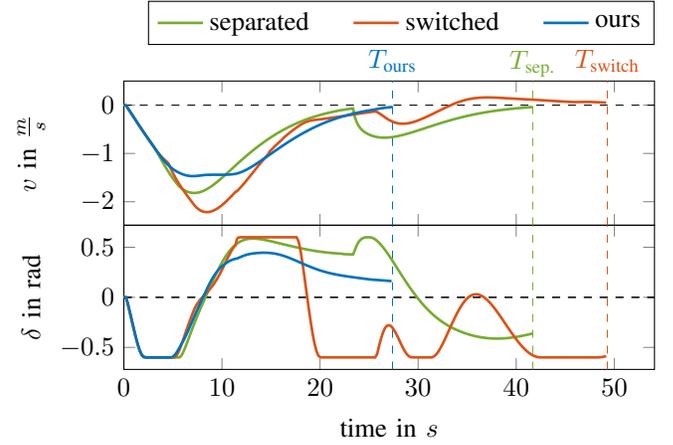

This behavior can be further analyzed in Fig.~\ref{fig:eval:dyn_weight_velocity}.
Here, it can be seen that our method reaches the goal earlier in a fluent movement without stopping or direction changes. 
Further, the maximum applied steering angle is smaller.
These trajectory details are further evaluated in Table~\ref{tab:traj_comparison_1}. 
Here, the trajectories are compared quantitatively with different metrics adapted from the CommonRoad benchmarks \cite{althoff2017}.
We evaluate the scenarios by calculating the Root Mean Square (RMS) values of various states and inputs of the trajectory.
We compare the RMS values of the steering angle $\delta_\text{RMS}$, the change of the steering angle $\dot\delta_\text{RMS}$, the longitudinal acceleration $a^\perp_\text{RMS}$, and lateral acceleration $a^\parallel_\text{RMS}$.
Low values indicate a smooth trajectory without excessive input changes or high acceleration values that can decrease passenger comfort.
Further, steering angles smaller than the maximum possible angle are also important to allow the underlying low-level controller to follow the trajectory, which increases the overall system's robustness.
Our approach outperforms the baseline approaches in all metrics except for the \emph{separated} approach in which they share the same RMS of the change of the steering angle $\dot\delta_\text{RMS}$.
Further, all trajectories reach the goal pose without collision.
\begin{table}[b]
    \caption{Scenario~1: Goal within the corridor}
        \vspace{-2mm}
	\label{tab:traj_comparison_1}
    \vspace{-0.5mm}
	\begin{center}
		\begin{tabular}{c | c | c | c | c | c | c}
			\toprule
			method & T & $\delta_\text{RMS}$ & $\dot\delta_\text{RMS}$ & $a^\parallel_\text{RMS}$& $a^\perp_\text{RMS}$ & safe \\
            \midrule
            ours & \textbf{27.39} & \textbf{0.018} & \textbf{0.00021} & \textbf{0.00038} & \textbf{0.0012} & \textbf{yes} \\ 
            separated& 41.68 & 0.037 & \textbf{0.00021} & 0.00063 & 0.0015 & \textbf{yes} \\ 
            switched & 49.29 & 0.059 & 0.00095 & 0.00051 & 0.0024 & \textbf{yes} \\ 
            \bottomrule
		\end{tabular}
	\end{center}
\end{table}%
In addition to that, the times at which the goal pose was reached can be observed.
Our approach reached the goal pose much earlier, with $T_\text{ours}\approx\qty{27}{\second}$, compared to $T_\text{sep.}\approx\qty{41}{\second}$ and $T_\text{switch}\approx\qty{49}{\second}$.
These evaluations and all the following ones can also be observed in the 
video\footnote{\label{fn:video}\href{https://youtu.be/28X5zaHW6bs}{https://youtu.be/28X5zaHW6bs}} provided with this paper.
\subsection{Dynamic Objective Allocation}\label{sec:eval:dyn_cost_type}
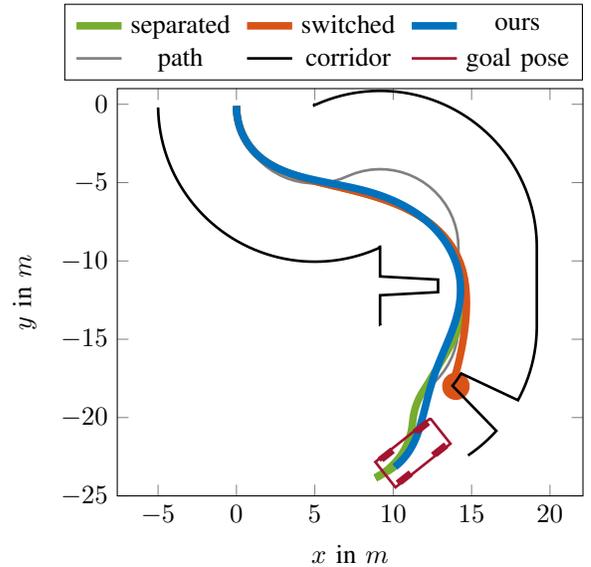
\begin{figure}[H]
    \centering
    \vspace{2mm}
    \input{plots/evaluation/dyn_cost/dyn_cost}
    \caption{%
    Scenario 2: The goal pose lies behind the corridor end. 
    Further, the corridor is narrowed by objects near the end.
    Our approach reaches the goal successfully while keeping distance to the corridor bounds, whereas the \emph{switched} approach collides with the boundary marked by the orange circle.
    The \emph{separated} approach also reaches the goal but needs an additional direction change.
    }%
    \label{fig:eval:dyn_cost}
\end{figure}

\begin{figure}[t]
    \centering
    \vspace{2mm}
    \input{plots/evaluation/dyn_cost/dyn_cost_velocity}%
    \caption{%
    Scenario 2: Driven velocities and steering angles of the scene in Fig.~\ref{fig:eval:dyn_cost}, in which the goal pose lies outside of the corridor.
    The \emph{separated} approach takes the longest time, in contrast to our method.
    The \emph{switched} baseline is not shown as it collided and thus is not a valid baseline.
    }%
    \label{fig:eval:dyn_cost_velocity}
\end{figure}
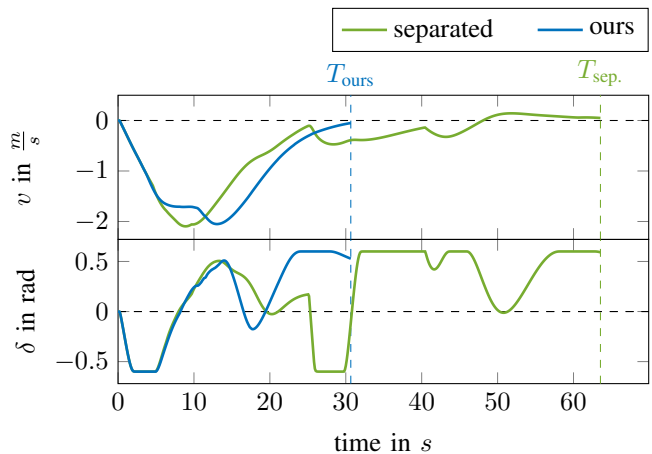

This section evaluates the dynamic objective allocation, which becomes relevant if a goal pose must be reached that is not longitudinally inside the corridor. 
This is called Scenario~2 in the following.
In this case, the corridor boundaries were also modified to emulate a narrow environment close to the goal pose.
Fig.~\ref{fig:eval:dyn_cost} visualizes the driven trajectories of the three different methods.
Here, our method reaches the goal pose outside of the corridor smoothly while keeping its distance to the boundaries of the corridor.
The \emph{separated} method also reaches the goal pose with a comparable driven trajectory.
However, it needs an additional direction change because the path end was not in an optimal position to plan to the goal pose.
In contrast to these methods, the method using the \emph{switched} baseline even collides with the corridor boundary near the end of the path.
This is because if the MPCC is completely switched to a Cartesian MPC, the Frenet coordinates are no longer valid.
Hence, all costs and constraints that are based on the Frenet coordinates are not calculated correctly, leading to the collision.
To summarize, the problem is to define when it is safe to switch to the Cartesian MPC.
Our proposed approach solves this problem inherently as the objective of each trajectory state is changed separately.
The advantages of our approach can also be observed in Fig.~\ref{fig:eval:dyn_cost_velocity}.
Here, the impact of the forced stop at the path end and the needed direction switches can be observed, leading to the \emph{separated} baseline needing approximately double of the time.
The three approaches are also compared quantitatively in Table~\ref{tab:traj_comparison_2}.
Our approach has the lowest $\delta_\text{RMS}$ and $\dot\delta_\text{RMS}$, while the \emph{separated} baseline has the lowest $a^\perp_\text{RMS}$ and $a^\parallel_\text{RMS}$.
In our approach, increasing the respective weights of the MPC could further lower the RMS values of both accelerations.
Because our approach reaches the goal way earlier than the \emph{separated} approach with $T_\text{ours}\approx\qty{30}{\second}$, compared to $T_\text{sep.}\approx\qty{63}{\second}$, the accelerations could thus be penalized more.
However, as already mentioned, the parameters are equal for all methods to allow for a fair comparison.
To summarize, our approach profits from the benefits of both baseline methods. It stays safe inside the corridor by keeping valid Frenet coordinates and plans inherently to the goal pose by switching the objective of all states behind the corridor to a pure cartesian one. 
This allows for the generation of a smooth and feasible trajectory.
\begin{table}[b]
    \vspace{2mm}
    \caption{Scenario 2: Goal outside the corridor}
    \vspace{-2mm}
	\label{tab:traj_comparison_2}
    \vspace{-0.5mm}
	\begin{center}
		\begin{tabular}{c | c | c | c | c | c | c}
			\toprule
			method & T & $\delta_\text{RMS}$ & $\dot\delta_\text{RMS}$ & $a^\parallel_\text{RMS}$& $a^\perp_\text{RMS}$ & safe \\
            \midrule
            ours & \textbf{30.66} & \textbf{0.035} & \textbf{0.00045} & 0.00072 & 0.0087 & \textbf{yes} \\ 
            separated & 63.55 & 0.043 & 0.0012 & \textbf{0.00024} & \textbf{0.0010} & \textbf{yes} \\ 
            switched & \sout{31.38} & \sout{0.011} & \sout{0.00020} & \sout{0.00075} & \sout{0.0045} & \textcolor{red}{no} \\ 
			\bottomrule
		\end{tabular}
	\end{center}
\end{table}%

\subsection{Evaluation of a docking maneuver in CARLA}
This section evaluates the proposed approach on the docking maneuver of the U-Shift II concept vehicle~\cite{ushift2}. 
It is an automated vehicle that can connect to different kinds of capsules, such as cargo capsules and public transport capsules.
This was done to separate the vehicle platform from its use case, which increases the flexibility of the mobility concept.
The vehicle can, for example, serve as a support vehicle for local bus fleets during rush hours and transport goods during the night, when few public transport vehicles are necessary.
For capsule swapping, the vehicle must detect these capsules and dock to them.
This is done by the pose estimation of so-called ChArUco boards~\cite{garrido2014charuco}.
More details about the automation of this vehicle can be found in ~\cite{buchholz2025ushift}.

\begin{figure}[t]
\centering
\vspace{2mm}
  \subfloat[Scene in CARLA]{
    \frame{%
        \includegraphics[
            trim={68cm 21cm 10cm 3cm},clip,
            width=0.95\columnwidth]{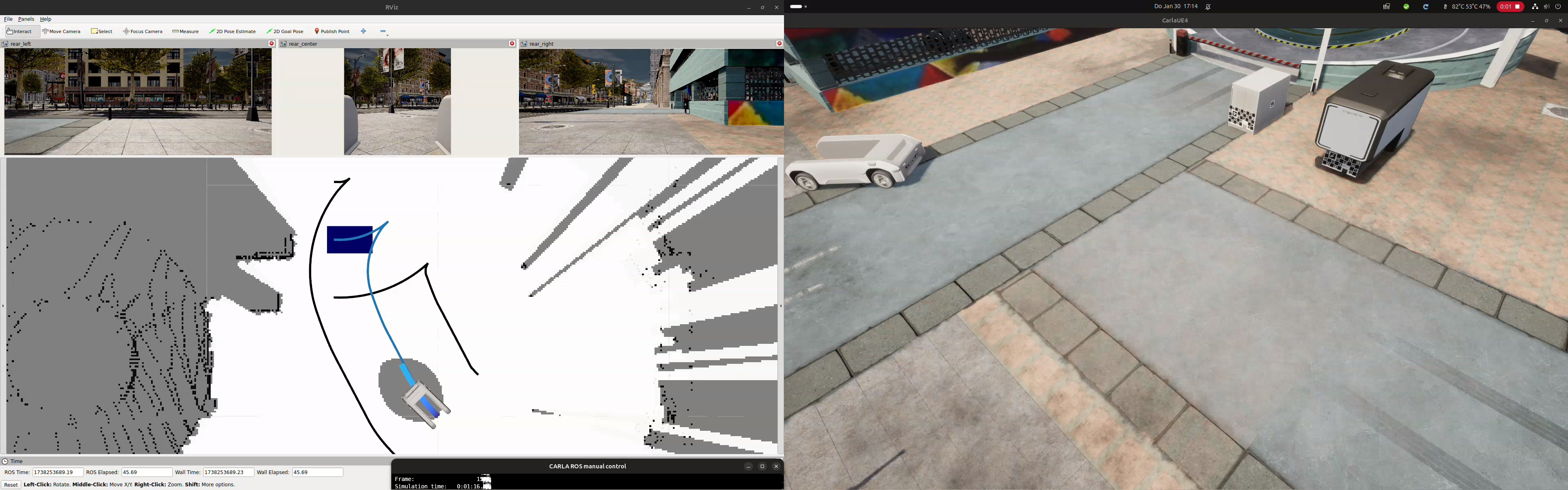}}
  }%
  \hfill
  \vspace{-2mm}
  \subfloat[Rviz view]{
    \frame{%
        \includegraphics[
            trim={10cm 5cm 77cm 18cm},clip,
            width=0.95\columnwidth]{img/evaluation/rviz_carla_overview.png}}
  }
  \caption{%
    Illustration of our software-in-the-loop test in CARLA.
    a) U-Shift vehicle on the left with two capsules on the right.
    The goal is to pickup the larger right one.
    b) The planned path and its corresponding trajectory on top of a grid map in Rviz~\cite{macenski2022ros}.
    The driving corridor is denoted in black.
    The path planning algorithm plans to the estimated docking position of the capsule because the capsule's actual pose has not been measured yet. 
    As soon as the capsule is detected, the estimated pose is passed to the dynamic objective MPC.
  }%
    \label{fig:eval:carla_scene}%
\end{figure}

\begin{figure}[t]
    \centering
    \input{plots/evaluation/carla_test/carla_test_velocity}%
    \caption{%
    Trajectory details of the docking maneuver in CARLA.
    Our approach generates a smooth trajectory to the goal pose at which the vehicle is lowered at $T_\text{goal}$.
    This takes some time, after which the vehicle reverses and connects with the capsule.
    }%
    \label{fig:eval:carla}
\end{figure}
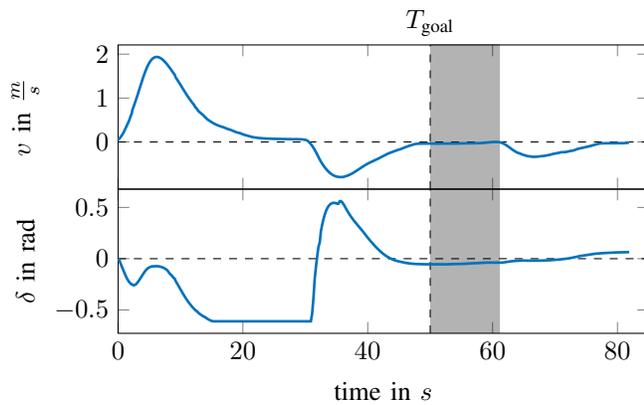

Hence, this section evaluates our proposed approach on the docking maneuver to a capsule.
The evaluation was done within the CARLA simulator~\cite{carla2017} with a reduced set of sensors. 
In this case, only three cameras facing to the rear and two lidars were simulated.
Fig.~\ref{fig:eval:carla_scene} shows the initial scene.
The goal pose for the proposed planning approach, is derived from the detected ChArUco board. 
The vehicle must reach this pose, being perfectly aligned right in front of the capsule to allow for successful docking.
Then, the chassis of the vehicle is lowered for capsule pick-up, leading to very limited steering capabilities. 
Thus, almost only straight driving is possible. 
Because this process is difficult to visualize by figures only, the docking maneuver is also visualized under the already referenced link\textsuperscript{\ref{fn:video}}.
The driven velocities and steering angles of the vehicle are shown in Fig.~\ref{fig:eval:carla}.
It can be observed that the vehicle drives forward and stops at the direction switch. 
Here, the cameras begin to detect the ChArUco board, which generates a dedicated goal pose for the planning algorithm.
Hence, the vehicle continues to follow the path and seamlessly reaches the goal pose.
At $T_\text{goal}\approx\qty{50}{\second}$, it stops to lower its chassis, which takes $\approx \qty{10}{\second}$, reverses slowly to its final position and attaches the capsule.

%% file: plots/evaluation/dyn_weight/dyn_weight.tex
\begin{tikzpicture}
\pgfplotstableread[col sep=comma,]{plots/evaluation/dyn_weight/new_lat_blend_driven.csv}\latblendeddriventable
\pgfplotstableread[col sep=comma,]{plots/evaluation/dyn_weight/new_lat_1_driven.csv}\latonedriventable
\pgfplotstableread[col sep=comma,]{plots/evaluation/dyn_weight/new_lat_50_driven.csv}\latfiftydriventable
\pgfplotstableread[col sep=comma,]{plots/evaluation/dyn_weight/new_lat_100_driven.csv}\lathundreddriventable
\pgfplotstableread[col sep=comma,]{plots/evaluation/dyn_weight/new_lat_blend_corridor.csv}\corridortable

\begin{axis}[
    width=0.9\linewidth, 
    height=7cm,
    axis equal,
    xlabel=$x$ in $m$,
    ylabel=$y$ in $m$,
    legend columns=3,
    legend style={at={(1.0, 1.02)},anchor=south east, /tikz/every even column/.append style={column sep=0.5cm}},
    xmin=-10,
    xmax=20,
    ymin=-0.5,
    ymax=28.0,
    clip marker paths=true,
]

\addplot[
    color=newGray, 
    line width=1pt,
    each nth point=5,
] 
table [
    x=path_x, 
    y=path_y, 
] {\corridortable};
\addlegendentry{path}

\addplot[
    color=black, 
    line width=1pt,
    each nth point=5,
] 
table [
    x=smooth_x_bound_left, 
    y=smooth_y_bound_left, 
] {\corridortable};
\addlegendentry{corridor}

\addplot[
    color=black, 
    line width=1pt,
    color=black, 
    each nth point=5,
    forget plot
] 
table [
    x=smooth_x_bound_right, 
    y=smooth_y_bound_right, 
] {\corridortable};



\addplot[
    color=matlabBlue,
    line width=1.2mm,
    each nth point = 2,
] 
table [
    x=sim_x, 
    y=sim_y,
] {\latblendeddriventable};
\addlegendentry{ours}

\addplot[
    color=matlabRed,
    line width=0.5mm,
    each nth point = 2,
] 
table [
    x=sim_x, 
    y=sim_y,
] {\latonedriventable};
\addlegendentry{$q^\text{c}=1$}

\addplot[
    color=matlabYellow,
    line width=0.5mm,
    each nth point = 2,
] 
table [
    x=sim_x, 
    y=sim_y,
] {\latfiftydriventable};
\addlegendentry{$q^\text{c}=50$}

\addplot[
    color=matlabGreen,
    line width=0.5mm,
    each nth point = 2,
] 
table [
    x=sim_x, 
    y=sim_y,
] {\lathundreddriventable};
\addlegendentry{$q^\text{c}=100$}

\end{axis}
\end{tikzpicture}

%% file: plots/evaluation/dyn_weight_goal/dyn_weight_goal.tex
\pgfplotstableread[col sep=comma,]{plots/evaluation/dyn_weight_goal/dyn_weight_driven.csv}\driventable
\pgfplotstableread[col sep=comma,]{plots/evaluation/dyn_weight_goal/dyn_weight_separate_driven.csv}\separatedriventable
\pgfplotstableread[col sep=comma,]{plots/evaluation/dyn_weight_goal/dyn_weight_switch_driven.csv}\hardswitchdriventable

\pgfplotstableread[col sep=comma,]{plots/evaluation/dyn_weight_goal/dyn_weight_corridor.csv}\corridortable
\pgfplotstableread[col sep=comma,]{plots/evaluation/dyn_weight_goal/dyn_weight_goal_pose.csv}\goaltable

\begin{tikzpicture}
    \begin{axis}[
        width=0.9\linewidth,
        height=6cm,
        axis equal,
        clip marker paths=true,
        ylabel=$y$ in $m$,
        xlabel=$x$ in $m$,
        xmin=-5.5,
        xmax=20,
        ymin=-17,
        ymax=0.5,
        legend columns = 3,
        legend style={at={(1.0, 1.02)},anchor=south east, /tikz/every even column/.append style={column sep=0.1cm}},
    ]
    \addplot[
    color=newGray, 
    line width=1pt,
    each nth point=5,
    forget plot,
    ] 
    table [
        x=path_x, 
        y=path_y, 
    ] {\corridortable};
    
    \addplot[
        color=black, 
        line width=1pt,
        color=black, 
        each nth point=5,
        forget plot
    ] 
    table [
        x=x_bound_right, 
        y=y_bound_right, 
    ] {\corridortable};

    \addplot[
        color = matlabGreen,
        line width=1.0mm,
        each nth point = 2,
    ] 
    table [
        x=sim_x, 
        y=sim_y,
    ] {\separatedriventable};
    \addlegendentry{separated}

        \addplot[
        color = matlabRed,
        line width=1.0mm,
        each nth point = 2,
    ] 
    table [
        x=sim_x, 
        y=sim_y,
    ] {\hardswitchdriventable};
    \addlegendentry{switched}

    \addplot[
        color = matlabBlue,
        line width=1.0mm,
        each nth point = 2,
    ] 
    table [
        x=sim_x, 
        y=sim_y,
    ] {\driventable};
    \addlegendentry{ours}

    \addlegendentry{path}
    \addlegendimage{no markers,     line width=1pt, newGray}

    \addplot[
    color=black, 
    line width=1pt,
    color=black,
    each nth point=5,
    ] 
    table [
        x=x_bound_left, 
        y=y_bound_left, 
    ] {\corridortable};
    \addlegendentry{corridor}
    
    \addplot[
        line width=1pt,
        color=matlabDarkRed,
        forget plot,
    ] 
    table [x=x, y=y] {\goaltable};
    \addlegendimage{area legend, matlabDarkRed}
    \addlegendentry{goal pose}
    
    \addplot[
        line width=1.5pt,
        color=matlabDarkRed
    ] 
    table [x=rl_x, y=rl_y] {\goaltable};
    \addplot[
        line width=1.5pt,
        color=matlabDarkRed
    ] 
    table [x=rr_x, y=rr_y] {\goaltable};
    \addplot[
        line width=1.5pt,
        color=matlabDarkRed
    ] 
    table [x=fl_x, y=fl_y] {\goaltable};
    \addplot[
        line width=1.5pt,
        color=matlabDarkRed
    ] 
    table [x=fr_x, y=fr_y] {\goaltable};

    \end{axis}
    \end{tikzpicture}

%% file: plots/evaluation/dyn_weight_goal/dyn_weight_velocity.tex

\pgfplotstableread[col sep=comma,]{plots/evaluation/dyn_weight_goal/dyn_weight_driven.csv}\driventable
\pgfplotstableread[col sep=comma,]{plots/evaluation/dyn_weight_goal/dyn_weight_separate_driven.csv}\separatedriventable
\pgfplotstableread[col sep=comma,]{plots/evaluation/dyn_weight_goal/dyn_weight_switch_driven.csv}\hardswitchdriventable

\def\Tours{27.390000000001482}
\def\Tseparate{41.6800000000002}
\def\Tswitch{49.28999999999876}
    
\begin{tikzpicture}
    \begin{groupplot}[
        group style={
            group size=1 by 2,
            x descriptions at=edge bottom,
            vertical sep=0pt,
        },
    ]

    \nextgroupplot[
    width=1.0\linewidth,
    height=3.5cm,
    ylabel=$v$ in $\frac{m}{s}$,
    xmin=0,
    ymax=0.5,
    legend columns = 3,
    legend style={at={(1.0, 1.25)},anchor=south east, /tikz/every even column/.append style={column sep=0.5cm}},
    clip=false,
    ]

    \node [text=matlabBlue, anchor=south] at 
    (axis cs: \Tours, \pgfkeysvalueof{/pgfplots/ymax}) {$T_\text{ours}$};
    \node [text=matlabRed, anchor=south] at 
    (axis cs:\Tswitch, \pgfkeysvalueof{/pgfplots/ymax}) {$T_\text{switch}$};
    \node [text=matlabGreen, anchor=south] at 
    (axis cs:\Tseparate, \pgfkeysvalueof{/pgfplots/ymax}-0.1) {$T_\text{sep.}$};
           
    \draw[dashed] 
    ({rel axis cs:0,0}|-{axis cs:0,0}) -- 
    ({rel axis cs:1,0}|-{axis cs:0,0}); 
    \draw[color=matlabBlue, dashed] 
    (axis cs:\Tours,\pgfkeysvalueof{/pgfplots/ymin})  -- 
    (axis cs:\Tours,\pgfkeysvalueof{/pgfplots/ymax});
    \draw[color=matlabGreen, dashed] 
    (axis cs:\Tseparate,\pgfkeysvalueof{/pgfplots/ymin}) -- 
    (axis cs:\Tseparate,\pgfkeysvalueof{/pgfplots/ymax});
    \draw[color=matlabRed, dashed] 
    (axis cs:\Tswitch,\pgfkeysvalueof{/pgfplots/ymin}) -- 
    (axis cs:\Tswitch,\pgfkeysvalueof{/pgfplots/ymax});

    \addplot[
        color=matlabGreen, 
        line width=1pt,
    ] 
    table [
        x=sim_t, 
        y=sim_v, 
    ] {\separatedriventable};
    \addlegendentry{separated}

    \addplot[
    color=matlabRed, 
    line width=1pt,
    each nth point = 2,
    ] 
    table [
        x=sim_t, 
        y=sim_v, 
    ] {\hardswitchdriventable};
    \addlegendentry{switched}
    
    \addplot[
        color=matlabBlue, 
        line width=1pt,
    ] 
    table [
        x=sim_t, 
        y=sim_v, 
    ] {\driventable};
    \addlegendentry{ours}

    \nextgroupplot[
        width=1.0\linewidth,
        height=3.5cm,
        xlabel=time in $s$,
        ylabel=$\delta$ in rad,
        xmin=0,
    ]
    \draw[dashed] ({rel axis cs:0,0}|-{axis cs:0,0}) -- ({rel axis cs:1,0}|-{axis cs:0,0});
    
    \draw[dashed] 
    ({rel axis cs:0,0}|-{axis cs:0,0}) -- 
    ({rel axis cs:1,0}|-{axis cs:0,0}); 
    \draw[color=matlabBlue, dashed] 
    (axis cs:\Tours,\pgfkeysvalueof{/pgfplots/ymin})  -- 
    (axis cs:\Tours,\pgfkeysvalueof{/pgfplots/ymax});
    \draw[color=matlabGreen, dashed] 
    (axis cs:\Tseparate,\pgfkeysvalueof{/pgfplots/ymin}) -- 
    (axis cs:\Tseparate,\pgfkeysvalueof{/pgfplots/ymax});
    \draw[color=matlabRed, dashed] 
    (axis cs:\Tswitch,\pgfkeysvalueof{/pgfplots/ymin}) -- 
    (axis cs:\Tswitch,\pgfkeysvalueof{/pgfplots/ymax});
    
    \addplot[
        color=matlabGreen, 
        line width=1pt,
        each nth point = 2,
    ] 
    table [
        x=sim_t, 
        y=sim_delta, 
    ] {\separatedriventable};

    \addplot[
    color=matlabRed, 
    line width=1pt,
    each nth point = 2,
    ] 
    table [
        x=sim_t, 
        y=sim_delta, 
    ] {\hardswitchdriventable};
    
    \addplot[
        color=matlabBlue, 
        line width=1pt,
        each nth point = 2,
    ] 
    table [
        x=sim_t, 
        y=sim_delta, 
    ] {\driventable};
    \end{groupplot}

\end{tikzpicture}

%% file: plots/evaluation/dyn_cost/dyn_cost.tex
\pgfplotstableread[col sep=comma,]{plots/evaluation/dyn_cost/dyn_cost_driven.csv}\driventable
\pgfplotstableread[col sep=comma,]{plots/evaluation/dyn_cost/dyn_cost_separate_driven.csv}\separatedriventable
\pgfplotstableread[col sep=comma,]{plots/evaluation/dyn_cost/dyn_cost_hard_switch_driven.csv}\hardswitchdriventable

\pgfplotstableread[col sep=comma,]{plots/evaluation/dyn_cost/dyn_cost_corridor.csv}\corridortable
\pgfplotstableread[col sep=comma,]{plots/evaluation/dyn_cost/dyn_cost_goal_pose.csv}\goaltable

\pgfplotsset{select coords between index/.style 2 args={
    x filter/.code={
        \ifnum\coordindex<#1\def\pgfmathresult{}\fi
        \ifnum\coordindex>#2\def\pgfmathresult{}\fi
    }
}}

\begin{tikzpicture}
    \begin{axis}[
        width=0.9\linewidth,
        height=7cm,
        axis equal,
        clip marker paths=true,
        ylabel=$y$ in $m$,
        xlabel=$x$ in $m$,
        xmin=-5.5,
        xmax=20,
        ymin=-25,
        ymax=1,
        legend columns = 3,
        legend style={at={(1.0, 1.02)},anchor=south east, /tikz/every even column/.append style={column sep=0.1cm}},
    ]

    \addplot[
    color=newGray, 
    line width=1pt,
    each nth point=5,
    forget plot,
    ] 
    table [
        x=path_x, 
        y=path_y, 
    ] {\corridortable};
    
    \addplot[
        color = matlabGreen,
        line width=1.0mm,
        each nth point = 2,
    ] 
    table [
        x=sim_x, 
        y=sim_y,
    ] {\separatedriventable};
    \addlegendentry{separated}

        \addplot[
        color = matlabRed,
        line width=1.0mm,
        each nth point = 2,
        select coords between index={0}{475}
    ] 
    table [
        x=sim_x, 
        y=sim_y,
    ] {\hardswitchdriventable};
    \addlegendentry{switched}

    \filldraw[matlabRed] (14,-18) circle (5pt) node[anchor=center]{};
 
    \addplot[
        color = matlabBlue,
        line width=1.0mm,
        each nth point = 2,
    ] 
    table [
        x=sim_x, 
        y=sim_y,
    ] {\driventable};
    \addlegendentry{ours}

    \addlegendentry{path}
    \addlegendimage{no markers,     line width=1pt, newGray}

    \addplot[
    color=black, 
    line width=1pt,
    color=black,
    each nth point=2,
    ] 
    table [
        x=x_bound_left, 
        y=y_bound_left, 
    ] {\corridortable};
    \addlegendentry{corridor}
    
    \addplot[
        line width=1pt,
        color=matlabDarkRed
    ] 
    table [x=x, y=y] {\goaltable};
    \addlegendentry{goal pose}
    
    \addplot[
        line width=1.5pt,
        color=matlabDarkRed
    ] 
    table [x=rl_x, y=rl_y] {\goaltable};
    \addplot[
        line width=1.5pt,
        color=matlabDarkRed
    ] 
    table [x=rr_x, y=rr_y] {\goaltable};
    \addplot[
        line width=1.5pt,
        color=matlabDarkRed
    ] 
    table [x=fl_x, y=fl_y] {\goaltable};
    \addplot[
        line width=1.5pt,
        color=matlabDarkRed
    ] 
    table [x=fr_x, y=fr_y] {\goaltable};
    
    \addplot[
        color=black, 
        line width=1pt,
        color=black, 
        each nth point=2,
        forget plot
    ] 
    table [
        x=x_bound_right, 
        y=y_bound_right, 
    ] {\corridortable};
    
    \end{axis}
\end{tikzpicture}

%% file: plots/evaluation/dyn_cost/dyn_cost_velocity.tex

\pgfplotstableread[col sep=comma,]{plots/evaluation/dyn_cost/dyn_cost_driven.csv}\driventable
\pgfplotstableread[col sep=comma,]{plots/evaluation/dyn_cost/dyn_cost_separate_driven.csv}\separatedriventable
\pgfplotstableread[col sep=comma,]{plots/evaluation/dyn_cost/dyn_cost_hard_switch_driven.csv}\hardswitchdriventable




\def\Tours{30.65997076034546}
\def\Tswitch{31.389970064163208}
\def\Tseparate{63.54993939399719}
    
\begin{tikzpicture}
    \begin{groupplot}[
        group style={
            group size=1 by 2,
            x descriptions at=edge bottom,
            vertical sep=0pt,
        },
    ]

    \nextgroupplot[
    width=1.0\linewidth,
    height=3.5cm,
    ylabel=$v$ in $\frac{m}{s}$,
    xmin=0,
    ymax=0.5,
    legend columns = 3,
    legend style={at={(1.0, 1.3)},anchor=south east, /tikz/every even column/.append style={column sep=0.5cm}},
    clip=false,
    ]

    \node [text=matlabBlue, anchor=south] at 
    (axis cs: \Tours, \pgfkeysvalueof{/pgfplots/ymax}) {$T_\text{ours}$};
    \node [text=matlabGreen, anchor=south] at 
    (axis cs:\Tseparate, \pgfkeysvalueof{/pgfplots/ymax}) {$T_\text{sep.}$};
           
    \draw[dashed] 
    ({rel axis cs:0,0}|-{axis cs:0,0}) -- 
    ({rel axis cs:1,0}|-{axis cs:0,0}); 
    \draw[color=matlabBlue, dashed] 
    (axis cs:\Tours,\pgfkeysvalueof{/pgfplots/ymin})  -- 
    (axis cs:\Tours,\pgfkeysvalueof{/pgfplots/ymax});
    \draw[color=matlabGreen, dashed] 
    (axis cs:\Tseparate,\pgfkeysvalueof{/pgfplots/ymin}) -- 
    (axis cs:\Tseparate,\pgfkeysvalueof{/pgfplots/ymax});

    \addplot[
        color=matlabGreen, 
        line width=1pt,
    ] 
    table [
        x=sim_t, 
        y=sim_v, 
    ] {\separatedriventable};
    \addlegendentry{separated}

    
    \addplot[
        color=matlabBlue, 
        line width=1pt,
    ] 
    table [
        x=sim_t, 
        y=sim_v, 
    ] {\driventable};
    \addlegendentry{ours}

    \nextgroupplot[
        width=1.0\linewidth,
        height=3.5cm,
        xlabel=time in $s$,
        ylabel=$\delta$ in rad,
        xmin=0,
    ]
    \draw[dashed] ({rel axis cs:0,0}|-{axis cs:0,0}) -- ({rel axis cs:1,0}|-{axis cs:0,0});
    
    \draw[dashed] 
    ({rel axis cs:0,0}|-{axis cs:0,0}) -- 
    ({rel axis cs:1,0}|-{axis cs:0,0}); 
    \draw[color=matlabBlue, dashed] 
    (axis cs:\Tours,\pgfkeysvalueof{/pgfplots/ymin})  -- 
    (axis cs:\Tours,\pgfkeysvalueof{/pgfplots/ymax});
    \draw[color=matlabGreen, dashed] 
    (axis cs:\Tseparate,\pgfkeysvalueof{/pgfplots/ymin}) -- 
    (axis cs:\Tseparate,\pgfkeysvalueof{/pgfplots/ymax});
    
    \addplot[
        color=matlabGreen, 
        line width=1pt,
    ] 
    table [
        x=sim_t, 
        y=sim_delta, 
    ] {\separatedriventable};

    
    \addplot[
        color=matlabBlue, 
        line width=1pt,
    ] 
    table [
        x=sim_t, 
        y=sim_delta, 
    ] {\driventable};
    \end{groupplot}

\end{tikzpicture}

%% file: plots/evaluation/carla_test/carla_test_velocity.tex

\pgfplotstableread[col sep=comma,]{plots/evaluation/carla_test/cleaned_carla_test_driven.csv}\driventable

\def\Tdocking{50}
    
\begin{tikzpicture}
    \begin{groupplot}[
        group style={
            group size=1 by 2,
            x descriptions at=edge bottom,
            vertical sep=0pt,
        },
    ]

    \nextgroupplot[
    width=1.0\linewidth,
    height=3.5cm,
    ylabel=$v$ in $\frac{m}{s}$,
    ytick={0, 1, 2},
    xmin=0,
    xmax=85,
    legend columns = 3,
    legend style={at={(1.0, 1.25)},anchor=south east, /tikz/every even column/.append style={column sep=0.5cm}},
    clip=false,
    ]

    \node [text=black, anchor=south] at 
    (axis cs: \Tdocking, \pgfkeysvalueof{/pgfplots/ymax}) {$T_\text{goal}$};
           
    \draw[dashed] 
    ({rel axis cs:0,0}|-{axis cs:0,0}) -- 
    ({rel axis cs:1,0}|-{axis cs:0,0}); 
    \draw[color=black, dashed] 
    (axis cs:\Tdocking,\pgfkeysvalueof{/pgfplots/ymin})  -- 
    (axis cs:\Tdocking,\pgfkeysvalueof{/pgfplots/ymax});

    \draw[draw=none, fill=black, fill opacity=0.3] (\Tdocking,\pgfkeysvalueof{/pgfplots/ymax}) rectangle (\Tdocking+11,\pgfkeysvalueof{/pgfplots/ymin});
    
    \addplot[
        color=matlabBlue, 
        line width=1pt,
        each nth point=2,
    ] 
    table [
        x=sim_t, 
        y=sim_v, 
    ] {\driventable};

    \nextgroupplot[
        width=1.0\linewidth,
        height=3.5cm,
        xlabel=time in $s$,
        ylabel=$\delta$ in rad,
        xmin=0,
        xmax=85,
    ]
    \draw[dashed] ({rel axis cs:0,0}|-{axis cs:0,0}) -- ({rel axis cs:1,0}|-{axis cs:0,0});

    \draw[color=black, dashed] 
    (axis cs:\Tdocking,\pgfkeysvalueof{/pgfplots/ymin})  -- 
    (axis cs:\Tdocking,\pgfkeysvalueof{/pgfplots/ymax});

    \draw[draw=none, fill=black, fill opacity=0.3] (\Tdocking,\pgfkeysvalueof{/pgfplots/ymax}) rectangle (\Tdocking+11,\pgfkeysvalueof{/pgfplots/ymin});
    
    \addplot[
        color=matlabBlue, 
        line width=1pt,
        each nth point=2,
    ] 
    table [
        x=sim_t, 
        y=sim_delta, 
    ] {\driventable};
    
    \end{groupplot}
\end{tikzpicture}

%% file: doc/07-conclusion.tex
\section{Conclusion}\label{sec:conclusion}
In this paper, we introduced a new MPC-based planning approach called dynamic objective MPC for precise and seamless motion planning to different kinds of goals in narrow environments.

At first, we proposed an adapted dynamic weight allocation method to plan precisely to direction switches and path ends.
This was the fundamental approach for the main contribution, called dynamic objective allocation. 
This novel method transforms an MPCC into a pure Cartesian MPC by state-dependent weight changes, allowing precise, safe, and seamless motion planning to specific goal poses in the proximity of a path without the need to change the planning algorithms.
Our approach provides shorter times to reach the goal pose and smoother trajectories compared to the baseline approaches.
Further, this approach can be used for seamless docking maneuvers, as shown in a software-in-the-loop simulation in CARLA in which a docking maneuver of the U-Shift concept vehicle to a capsule was shown.

In future work, we want to investigate how the dynamic weight allocation and dynamic objective allocation methods affect the stability of the MPC and under what circumstances the stability of the approach can be assured.
At last, we want to apply this algorithm in real-world scenarios of the U-Shift concept vehicle, in which it will be used for normal trajectory planning and docking maneuvers.

%% file: root.bbl
\begin{thebibliography}{10}
\providecommand{\url}[1]{#1}
\csname url@rmstyle\endcsname
\providecommand{\newblock}{\relax}
\providecommand{\bibinfo}[2]{#2}
\providecommand\BIBentrySTDinterwordspacing{\spaceskip=0pt\relax}
\providecommand\BIBentryALTinterwordstretchfactor{4}
\providecommand\BIBentryALTinterwordspacing{\spaceskip=\fontdimen2\font plus
\BIBentryALTinterwordstretchfactor\fontdimen3\font minus
  \fontdimen4\font\relax}
\providecommand\BIBforeignlanguage[2]{{%
\expandafter\ifx\csname l@#1\endcsname\relax
\typeout{** WARNING: IEEEtran.bst: No hyphenation pattern has been}%
\typeout{** loaded for the language `#1'. Using the pattern for}%
\typeout{** the default language instead.}%
\else
\language=\csname l@#1\endcsname
\fi
#2}}

\bibitem{warehouseReport}
\BIBentryALTinterwordspacing
{ABI research}. (2019) 50,000 warehouses to use robots by 2025 as barriers to
  entry fall and ai innovation accelerates. [Online]. Available:
  \url{https://www.abiresearch.com/press/50000-warehouses-use-robots-2025-barriers-entry-fall-and-ai-innovation-accelerates}
\BIBentrySTDinterwordspacing

\bibitem{macenski2020marathon2}
S.~Macenski, F.~Martín, R.~White, and J.~G. Clavero, ``The marathon 2: A
  navigation system,'' in \emph{2020 IEEE/RSJ International Conference on
  Intelligent Robots and Systems (IROS)}, 2020, pp. 2718--2725.

\bibitem{macenski2023survey}
\BIBentryALTinterwordspacing
S.~Macenski, T.~Moore, D.~V. Lu, A.~Merzlyakov, and M.~Ferguson, ``From the
  desks of ros maintainers: A survey of modern {\&} capable mobile robotics
  algorithms in the robot operating system 2,'' \emph{Robotics and Autonomous
  Systems}, vol. 168, p. 104493, 2023. [Online]. Available:
  \url{https://www.sciencedirect.com/science/article/pii/S092188902300132X}
\BIBentrySTDinterwordspacing

\bibitem{macenski2022ros}
\BIBentryALTinterwordspacing
S.~Macenski, T.~Foote, B.~Gerkey, C.~Lalancette, and W.~Woodall, ``Robot
  operating system 2: Design, architecture, and uses in the wild,''
  \emph{Science Robotics}, vol.~7, no.~66, p. eabm6074, 2022. [Online].
  Available: \url{https://www.science.org/doi/abs/10.1126/scirobotics.abm6074}
\BIBentrySTDinterwordspacing

\bibitem{thrun2006stanley}
S.~Thrun, \emph{et~al.}, ``Stanley: The robot that won the darpa grand
  challenge,'' \emph{Journal of Field Robotics}, vol.~23, no.~9, pp. 661 --
  692, June 2006.

\bibitem{coulter1992pure}
R.~C. Coulter, ``Implementation of the pure pursuit path tracking algorithm,''
  Carnegie Mellon University, Pittsburgh, PA, Tech. Rep. CMU-RI-TR-92-01,
  January 1992.

\bibitem{macenski2023regulated}
\BIBentryALTinterwordspacing
S.~Macenski, S.~Singh, F.~Mart{\'i}n, and J.~Gin{\'e}s, ``Regulated pure
  pursuit for robot path tracking,'' \emph{Autonomous Robots}, vol.~47, no.~6,
  pp. 685--694, Aug 2023. [Online]. Available:
  \url{https://doi.org/10.1007/s10514-023-10097-6}
\BIBentrySTDinterwordspacing

\bibitem{fox1997dwa}
D.~Fox, W.~Burgard, and S.~Thrun, ``The dynamic window approach to collision
  avoidance,'' \emph{IEEE Robotics \& Automation Magazine}, vol.~4, no.~1, pp.
  23--33, 1997.

\bibitem{williams2016mppi}
G.~Williams, P.~Drews, B.~Goldfain, J.~M. Rehg, and E.~A. Theodorou,
  ``Aggressive driving with model predictive path integral control,'' in
  \emph{2016 IEEE International Conference on Robotics and Automation (ICRA)},
  2016, pp. 1433--1440.

\bibitem{ziegler2014bertha}
J.~Ziegler, P.~Bender, T.~Dang, and C.~Stiller, ``Trajectory planning for
  bertha — a local, continuous method,'' in \emph{2014 IEEE Intelligent
  Vehicles Symposium Proceedings}, 2014, pp. 450--457.

\bibitem{liniger2015}
\BIBentryALTinterwordspacing
A.~Liniger, A.~Domahidi, and M.~Morari, ``Optimization-based autonomous racing
  of 1:43 scale rc cars,'' \emph{Optimal Control Applications and Methods},
  vol.~36, no.~5, pp. 628--647, 2015. [Online]. Available:
  \url{https://onlinelibrary.wiley.com/doi/abs/10.1002/oca.2123}
\BIBentrySTDinterwordspacing

\bibitem{romero2022}
A.~Romero, S.~Sun, P.~Foehn, and D.~Scaramuzza, ``Model predictive contouring
  control for time-optimal quadrotor flight,'' \emph{IEEE Transactions on
  Robotics}, vol.~38, no.~6, pp. 3340--3356, 2022.

\bibitem{park2011graceful}
J.~J. Park and B.~Kuipers, ``A smooth control law for graceful motion of
  differential wheeled mobile robots in 2d environment,'' in \emph{2011 IEEE
  International Conference on Robotics and Automation (ICRA)}, 2011, pp.
  4896--4902.

\bibitem{fuchshumer2005flatness}
S.~Fuchshumer, K.~Schlacher, and T.~Rittenschober, ``Nonlinear vehicle dynamics
  control - a flatness based approach,'' in \emph{Proceedings of the 44th IEEE
  Conference on Decision and Control}, 2005, pp. 6492--6497.

\bibitem{chen2019ilqr}
J.~Chen, W.~Zhan, and M.~Tomizuka, ``Autonomous driving motion planning with
  constrained iterative lqr,'' \emph{IEEE Transactions on Intelligent
  Vehicles}, vol.~4, no.~2, pp. 244--254, 2019.

\bibitem{zhang2018opt}
X.~Zhang, A.~Liniger, A.~Sakai, and F.~Borrelli, ``Autonomous parking using
  optimization-based collision avoidance,'' in \emph{2018 IEEE Conference on
  Decision and Control (CDC)}, 2018, pp. 4327--4332.

\bibitem{dexory}
\BIBentryALTinterwordspacing
Dexory. (2024) The robotics behind dexory’s automated inventory management
  system. [Online]. Available:
  \url{https://www.dexory.com/insights/the-robotics-behind-dexorys-automated-inventory-management-system}
\BIBentrySTDinterwordspacing

\bibitem{acados2021}
R.~Verschueren, \emph{et~al.}, ``acados -- a modular open-source framework for
  fast embedded optimal control,'' \emph{Mathematical Programming Computation},
  2021.

\bibitem{althoff2017}
M.~Althoff, M.~Koschi, and S.~Manzinger, ``Commonroad: Composable benchmarks
  for motion planning on roads,'' in \emph{2017 IEEE Intelligent Vehicles
  Symposium (IV)}, 2017, pp. 719--726.

\bibitem{ushift2}
M.~Münster, \emph{et~al.}, ``{U-Shift II} vision and project goals,'' in
  \emph{22. Internationales Stuttgarter Symposium}, M.~Bargende, H.-C. Reuss,
  and A.~Wagner, Eds.\hskip 1em plus 0.5em minus 0.4em\relax Springer
  Fachmedien Wiesbaden, 2022, pp. 18--31.

\bibitem{garrido2014charuco}
\BIBentryALTinterwordspacing
S.~Garrido-Jurado, R.~Mu\~{n}oz Salinas, F.~Madrid-Cuevas, and
  M.~Mar\'{\i}n-Jim\'{e}nez, ``Automatic generation and detection of highly
  reliable fiducial markers under occlusion,'' \emph{Pattern Recogn.}, vol.~47,
  no.~6, p. 2280–2292, June 2014. [Online]. Available:
  \url{https://doi.org/10.1016/j.patcog.2014.01.005}
\BIBentrySTDinterwordspacing

\bibitem{buchholz2025ushift}
M.~Buchholz, T.~Wodtko, O.~Schumann, and D.~Authaler, ``The automation concept
  of the {U-Shift II} vehicle,'' \emph{accepted for publication at 2025
  Stuttgart International Symposium}, 2025.

\bibitem{carla2017}
A.~Dosovitskiy, G.~Ros, F.~Codevilla, A.~Lopez, and V.~Koltun, ``{CARLA}: {An}
  open urban driving simulator,'' in \emph{Proceedings of the 1st Annual
  Conference on Robot Learning}, 2017, pp. 1--16.

\end{thebibliography}
